\documentclass{article} 
\usepackage[final]{colm2026_conference}

\usepackage{CJKutf8}

\usepackage{hyperref}
\usepackage{url}

\usepackage{amssymb}
\usepackage{amsmath}
\usepackage{amsthm}
\usepackage{amsfonts}
\usepackage{bm}
\usepackage{nicefrac}
\usepackage{dsfont}
\usepackage{fancyvrb}
\usepackage{fvextra}
\usepackage{listings}
\usepackage{xcolor}
\usepackage{tabularx}
\usepackage[all]{nowidow}
\usepackage{xcolor}
\usepackage{colortbl}
\usepackage{changepage}  

\usepackage{booktabs}
\usepackage{multirow}
\usepackage{makecell}
\usepackage{arydshln}

\usepackage{comment}
\usepackage{xspace}
\usepackage{soul}

\usepackage{enumitem}
\usepackage[normalem]{ulem}
\setlist[itemize,enumerate]{leftmargin=*}

\usepackage{pgfplots}
\usepackage{tikz}
\usetikzlibrary{calc}
\pgfplotsset{compat=1.9}

\makeatletter
\def\adl@drawiv#1#2#3{%
        \hskip.5\tabcolsep
        \xleaders#3{#2.5\@tempdimb #1{1}#2.5\@tempdimb}%
                #2\z@ plus1fil minus1fil\relax
        \hskip.5\tabcolsep}
\newcommand{\cdashlinelr}[1]{%
  \noalign{\vskip 2pt
           \global\let\@dashdrawstore\adl@draw
           \global\let\adl@draw\adl@drawiv}
  \cdashline{#1}[.4pt/2pt]
  \noalign{\global\let\adl@draw\@dashdrawstore
           \vskip 2pt}}
\makeatother

\definecolor{light-orange}{HTML}{fee9d4}
\definecolor{light-green}{HTML}{d8f0d3}
\definecolor{light-blue}{HTML}{dae8f5}
\definecolor{set10-red}{HTML}{e41a1c}
\definecolor{set10-blue}{HTML}{377eb8}
\definecolor{set10-green}{HTML}{4daf4a}

\definecolor{CustomBlue}{RGB}{57,83,191}
\definecolor{CustomRed}{HTML}{a75151}
\definecolor{DarkGreenOne}{RGB}{106,168,79}

\definecolor{SwordOrange}{HTML}{ff8351}
\definecolor{SwordBlueComplentarySwordOrange}{HTML}{51cdff}
\definecolor{SwordBlue}{HTML}{5993ea}
\definecolor{SwordSilver}{HTML}{a4aab6}
\definecolor{SwordTan}{HTML}{dacdc3}
\definecolor{SwordNoir}{HTML}{20222c}
\definecolor{SwordRed}{HTML}{ff8283}
\definecolor{SwordYellow}{HTML}{ffda51}
\definecolor{SwordSquash}{HTML}{00c487}
\definecolor{SwordPink}{HTML}{ed75ff}

\definecolor{QwenPurple}{HTML}{6349ea}
\definecolor{GeminiBlue}{HTML}{4185f4}
\definecolor{OpenAIGreen}{HTML}{1fa681}
\definecolor{AnthropicTan}{HTML}{d5a583}
\definecolor{ZaiAsh}{HTML}{282828}
\usepackage[most]{tcolorbox}

\newtcbox{\clustertab}[1]{on line, box align=base, colback={#1},colframe={#1},size=fbox,arc=2pt,top=-1.5pt, bottom=-1.5pt, left=-1.5pt, right=-1.5pt, boxrule=0pt, enlarge left by=1pt}

\usepackage{pifont}
\newcommand{\spbci}[2]{\,{\textcolor{gray}{\tiny[#1,\,#2]}}}

\pgfplotsset{
  spb axis/.style={
    axis x line*=bottom,
    axis y line*=left,
    y axis line style={draw=black!40},
    x axis line style={draw=black!40},
    ymajorgrids,
    xmajorgrids,
    grid style={draw=gray!20},
    tick style={draw=black!40, thin},
    tick align=outside,
    legend style={
      draw=none,
      fill=none,
      legend cell align=left,
    },
  },
}

\usepackage{float}

\tcbuselibrary{listings}
\usepackage{fancyvrb}
\usepackage{microtype}
\usepackage{graphicx}
\usepackage{rotating}
\usepackage{float}

\usepackage{lineno}

\definecolor{darkblue}{rgb}{0, 0, 0.5}
\hypersetup{colorlinks=true, citecolor=darkblue, linkcolor=darkblue, urlcolor=darkblue}

\title{Self-Preference Bias in Rubric-Based Evaluation \\ of Large Language Models}

\author{
  José Pombal$^{1,2,3}$, Ricardo Rei$^{1}$ \& André F. T. Martins$^{2, 3, 4, 5}$ \\
  \ \\
  $^1$Sword Health, $^2$Instituto de Telecomunicações
  \\
  $^3$Instituto Superior Técnico, Universidade de Lisboa, $^4$TransPerfect, $^5$ELLIS Unit Lisbon
  \\
  \texttt{j.pombal@swordhealth.com}
}

\begin{document}

\ifcolmsubmission
\linenumbers
\fi

\maketitle

\begin{abstract}
LLM-as-a-judge has become the \textit{de facto} approach for evaluating LLM outputs. However, judges are known to exhibit \textit{self-preference bias} (SPB): they tend to favor outputs produced by themselves or by models from their own family.
This skews evaluations and, thus, hinders model development, especially in settings of recursive self-improvement.
We present the first study of SPB in rubric-based evaluation, an increasingly popular benchmarking paradigm where judges issue binary verdicts on individual evaluation criteria, instead of assigning holistic scores or rankings.
Using IFEval and LiveCodeBench, benchmarks with programmatically verifiable rubrics, we show that SPB persists even when evaluation criteria are entirely objective: among rubrics where generators fail, judges can be more than 50\% more likely to incorrectly mark them as satisfied when the output is their own.
We also find that, similarly to other evaluation paradigms, ensembling multiple judges helps mitigate SPB, but without fully eliminating it.
On HealthBench, a medical chat benchmark with subjective rubrics, we observe that SPB skews model scores by up to 10 points, a potentially decisive margin when ranking frontier models.
We analyze the factors that drive SPB in this setting, finding that negative rubrics and subjective topics like communication and emergency referrals are particularly susceptible.
\end{abstract}

\section{Introduction}\label{sec:introduction}

Automated evaluation via LLM-as-a-judge has become the standard practice for assessing model outputs~\citep{zheng2023judging}.
In this paradigm, an LLM is typically prompted to either compare two outputs and select the best one (pairwise comparison, PWC), or to rate an output in isolation according to a predefined numeric scale (direct assessment, DA).
However, among other issues, self-preference bias (SPB) can occur under both approaches, whereby judges show a systematic preference for, or attribute disproportionately higher scores to outputs produced by themselves, by models from their own family, or by models trained on data they generated or curated~\citep{verga2024replacing,dietz2025llm,li2025preference,spiliopoulou2025play,zhao2024language,pugachev2025repa,wataoka2024self,koo2023benchmarking}.
SPB may skew evaluations and misguide practitioners into believing certain systems are better than others.
In self-improvement pipelines, SPB can lead to a feedback loop where models are optimized to perform well according to their own preferences, rather than truly improving on the task at hand~\citep{xu2024pride,panickssery2024llm}.

\begin{figure}[t]
    \centering
    \includegraphics[width=\textwidth]{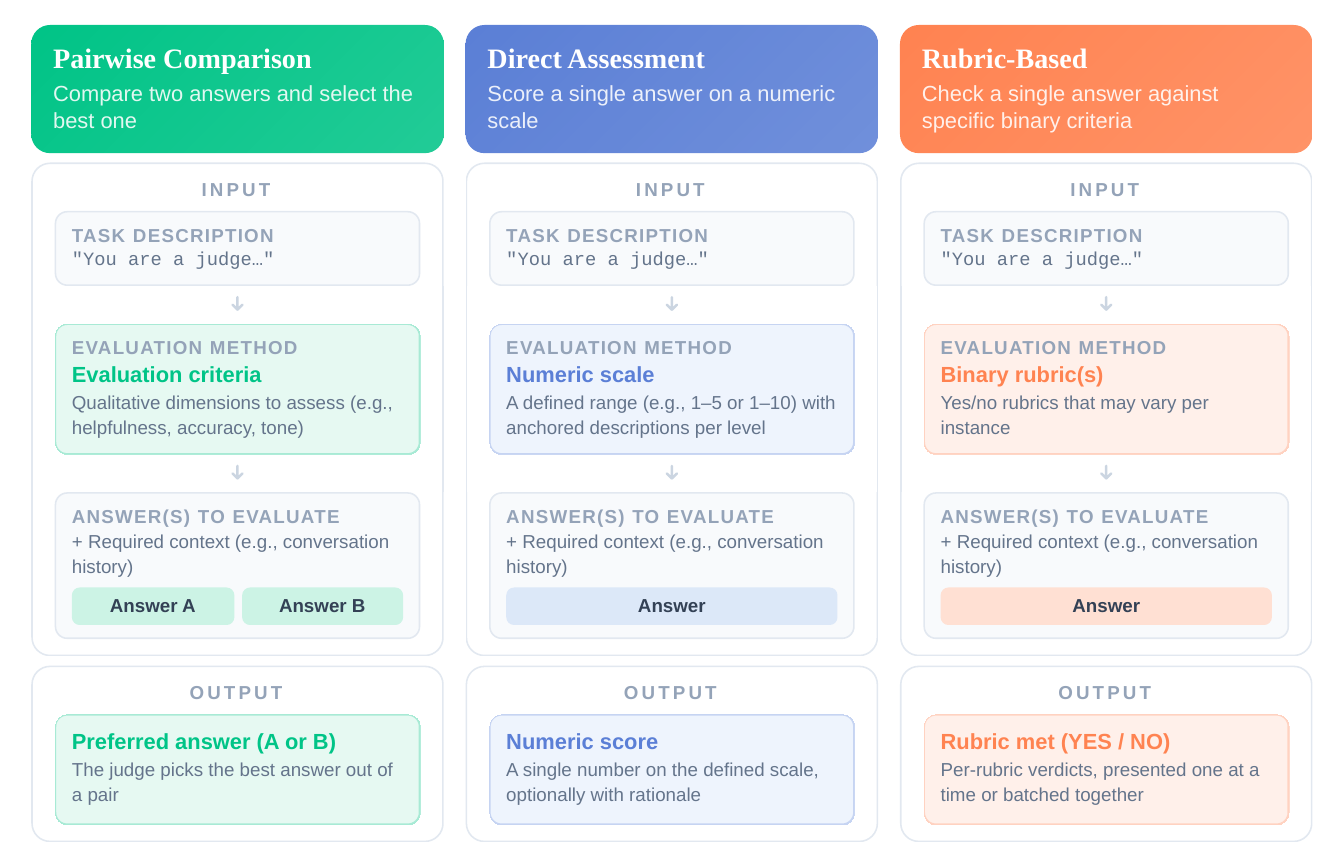}
    \caption{Illustration of three LLM-as-a-Judge paradigms: pairwise comparison (PWC), direct assessment (DA) and rubric-based evaluation (RB).}
    \label{fig:judge-paradigms}
\end{figure}

More recently, rubric-based evaluation (RB) has emerged as an increasingly popular approach for benchmarking~\citep{arora2025healthbench,starace2025paperbench,lin2024wildbench,sirdeshmukh2025multichallenge} and reward modeling in reinforcement learning~\citep{huang2025reinforcement,gunjal2025rubrics,dineen-etal-2025-qa}.
In this setting, the judge is presented with one or more binary rubrics an output should satisfy (see Figure~\ref{fig:judge-paradigms}).
Rubric-based evaluation allows for more granular and interpretable evaluations than PWC and DA, and has been shown to correlate well with human judgments~\citep{arora2025healthbench,starace2025paperbench,lin2024wildbench,sirdeshmukh2025multichallenge}.
Furthermore, by discarding the need for direct comparisons between outputs and for numeric scores, it eliminates the response ordering and score calibration issues found in PWC and DA, respectively.
For the same reasons, RB could also be less susceptible to SPB, though this has not been systematically studied.

In this work, we first show that judges do, in fact, exhibit systematic preference for their own outputs---or those of their model family---in rubric-based evaluation, even when rubrics are entirely objective (\textit{i.e.,} programmatically verifiable), and when controlling for instance difficulty and judge quality.
For outputs that fail to satisfy a given rubric, models can be more than 50\% more likely to incorrectly mark it as satisfied when the output was generated by themselves or their model family.
Furthermore, the extent of self-preference bias in rubric-based evaluation is similar to that of direct assessment but significantly less severe than that of pairwise comparison, and, while the popular approach of ensembling multiple judges to mitigate SPB also proves effective, it does not fully eliminate the bias.
We reach similar conclusions analyzing HealthBench~\citep{arora2025healthbench}, a realistic benchmark for medical chat assistant evaluation with subjective human-written rubrics.
GPT-5, for example, tends to overestimate its performance by about 4 points relative to an ensemble of judges, a difference that may prove substantial when ranking frontier models and making claims about their capabilities.

We analyze the factors that contribute to SPB in HealthBench, uncovering several actionable patterns: (1) negative rubrics, which penalize generators if fulfilled, are more vulnerable to SPB than positive ones, (2) rubrics targeting communication and emergency referrals, among other themes, elicit higher SPB than those assessing accuracy, (3) inter-judge agreement can serve as a practical signal for flagging high-SPB rubrics.
Together, these findings offer concrete guidance for designing rubrics that are more resilient to self-preference bias and point to promising directions for its mitigation.\footnote{We make code to reproduce our experiments and analysis available on \href{https://github.com/zeppombal/rubric-self-preference}{Github}.}

\section{Experimental setup}\label{sec:experimental-setup}

\subsection{Measuring self-preference bias}\label{subsec:measuring-spb}

Self-preference bias (SPB) is the tendency of judges to unjustly and systematically favor their own outputs.
To define a metric for SPB, we first need to define what \emph{unjustly favoring} means, since judges may have legitimate reasons to prefer their own outputs, or a general tendency to over- or under-estimate any generator.

Generically, we can characterize a judge's disposition toward a particular generator through its \emph{overestimation rate}: the proportion of cases in which the judge rules more favorably for a generator than the ground truth warrants.
Concretely, let $\mathcal{O}(\mathcal{J}, \mathcal{G})$ denote the non-negative overestimation rate of judge $\mathcal{J}$ toward generator $\mathcal{G}$, computed over a benchmark $\mathcal{D}$ (we define $\mathcal{O}$ precisely below for each evaluation paradigm in Figure~\ref{fig:judge-paradigms}).
Self-preference then corresponds to the overestimation rate when the generator is the judge itself, $\mathcal{O}(\mathcal{J}, \mathcal{J})$.
However, a high $\mathcal{O}(\mathcal{J}, \mathcal{J})$ alone does not constitute bias: the judge may simply be a lenient evaluator across the board.
The concern arises when the judge overestimates its own outputs \emph{disproportionately} relative to those of unrelated models.
We capture this through the \emph{Harmful Self-Preference Propensity Ratio} (HSPP Ratio, an extension of HSPP~\citep{chen2025llm}):
\begin{equation}\label{eq:hspp-ratio}
    \text{HSPP-R}_\text{self}(\mathcal{J}) = \frac{\mathcal{O}(\mathcal{J}, \mathcal{J})}{\frac{1}{|\mathcal{S}_\mathcal{J}|} \displaystyle\sum_{\mathcal{G} \in \mathcal{S}_\mathcal{J}} \mathcal{O}(\mathcal{J}, \mathcal{G})}\,,
\end{equation}
where $\mathcal{S}_\mathcal{J}$ is the set of generators that are not $\mathcal{J}$ and do not belong to the same model family as $\mathcal{J}$.
We exclude same-family generators from $\mathcal{S}_\mathcal{J}$ to prevent family-level biases from skewing self-preference.
An HSPP-R of 1 indicates no self-preference; values above 1 indicate the judge overestimates its own outputs more than those of unrelated models.

Since models within the same family are often similar, SPB may extend beyond the judge itself to its relatives.\footnote{We define ``relatives'' as models belonging to the same family (e.g., Gemma 3 27B and 12B). However, most models likely share some degree of \textit{relatedness}, due to distillation or similar training data. How to best define and measure \textit{relatedness} remains an open question.}
We therefore also define a \emph{family} variant:
\begin{equation}\label{eq:hspp-ratio-family}
    \text{HSPP-R}_\text{fam}(\mathcal{J}) = \frac{\frac{1}{|\mathcal{F}_\mathcal{J}|}\displaystyle\sum_{\mathcal{G} \in \mathcal{F}_\mathcal{J}} \mathcal{O}(\mathcal{J}, \mathcal{G})}{\frac{1}{|\mathcal{S}_\mathcal{J}|} \displaystyle\sum_{\mathcal{G} \in \mathcal{S}_\mathcal{J}} \mathcal{O}(\mathcal{J}, \mathcal{G})}\,,
\end{equation}
where $\mathcal{F}_\mathcal{J}$ is the set of generators from the same family as $\mathcal{J}$ (excluding $\mathcal{J}$ to isolate family-only effects), and the denominator is the same as in Equation~\ref{eq:hspp-ratio}. The set of all generators is the disjoint union $\{\mathcal{J}\} \cup \mathcal{F}_{\mathcal{J}} \cup \mathcal{S}_{\mathcal{J}}$.

\paragraph{Instance-level overestimation.}
For evaluation methods that produce a scalar score per instance, we compare the judge's instance-level orderings against the ground-truth orderings. This score may be derived from rubric satisfaction fractions (when the judge evaluates each rubric individually (single-rubric, SR), or all at once (all-rubrics, AR)), from a direct assessment scale (DA), or from pairwise comparison outcomes (PWC).
For each instance $x \in \mathcal{D}$, let $s_\mathcal{J}(\mathcal{G}, x)$ be the judge's score for generator $\mathcal{G}$ on instance $x$, and let $s^*(\mathcal{G}, x)$ be the corresponding ground-truth score.\footnote{For SR and AR, $s$ is the fraction of satisfied rubrics; for DA, we prompt the judge for the fraction of satisfied rubrics directly; for PWC, the resolved outcome is mapped to $\{+1, 0, -1\}$ for win, tie, loss.}
Let $\mathcal{G}' \neq \mathcal{G}$ be an opponent generator, and let the judge outcome be $w_{\mathcal{J}}(\mathcal{G}, \mathcal{G}', x) = \mathrm{sgn}\big(s_\mathcal{J}(\mathcal{G}, x) - s_\mathcal{J}(\mathcal{G}', x)\big)$ and the reference outcome $w^*(\mathcal{G}, \mathcal{G}', x) = \mathrm{sgn}\big(s^*(\mathcal{G}, x) - s^*(\mathcal{G}', x)\big)$, where $\mathrm{sgn}(a)$ is $+1$, $0$, or $-1$ according to whether $a$ is positive ($\mathcal{G}$ wins), zero (tie), or negative ($\mathcal{G}$ loses).
We define overestimation as the false-positive rate of the judge on instances where $\mathcal{G}$ should lose (\textit{i.e.,} $w^*(\mathcal{G}, \mathcal{G}', x) = -1$), measuring how often the judge nonetheless favors $\mathcal{G}$ (\textit{i.e.,} $w_{\mathcal{J}}(\mathcal{G}, \mathcal{G}', x) > w^*(\mathcal{G}, \mathcal{G}', x)$):
\begin{equation}\label{eq:overest-instance}
    \mathcal{O}_\text{inst}(\mathcal{J}, \mathcal{G}) = \frac{\displaystyle\sum_{\mathcal{G}' \neq \mathcal{G}} \sum_{x \in \mathcal{D}} \mathbb{1}\!\left\{w_{\mathcal{J}}(\mathcal{G}, \mathcal{G}', x) > w^*(\mathcal{G}, \mathcal{G}', x),\; w^*(\mathcal{G}, \mathcal{G}', x) = -1\right\}}{\displaystyle\sum_{\mathcal{G}' \neq \mathcal{G}} \sum_{x \in \mathcal{D}} \mathbb{1}\!\left\{w^*(\mathcal{G}, \mathcal{G}', x) = -1\right\}}\,,
\end{equation}

\paragraph{Rubric-level overestimation.}
For methods that produce individual rubric judgments (SR and AR), we can measure overestimation at a finer granularity.
Let $b_\mathcal{J}(\mathcal{G}, x, k) \in \{-1, 1\}$ be the judge's verdict on rubric $k$ for generator $\mathcal{G}$ on instance $x$, and $b^*(\mathcal{G}, x, k)$ the ground-truth verdict.
A rubric-level overestimation occurs when the judge marks a rubric as satisfied ($b_\mathcal{J} = 1$) while the ground truth marks it as unsatisfied ($b^* = -1$):
\begin{equation}\label{eq:overest-rubric}
    \mathcal{O}_\text{rub}(\mathcal{J}, \mathcal{G}) = \frac{\displaystyle\sum_{x \in \mathcal{D}} \sum_{k} \mathbb{1}\big\{b_\mathcal{J}(\mathcal{G}, x, k) = 1 \;\wedge\; b^*(\mathcal{G}, x, k) = -1\big\}}{\displaystyle\sum_{x \in \mathcal{D}} \sum_{k} \mathbb{1}\big\{b^*(\mathcal{G}, x, k) = -1\big\}}\,.
\end{equation}
This is the false positive rate of the judge restricted to the outputs of a single generator and can be interpreted as: among rubrics that a generator objectively fails, how often does the judge incorrectly mark them as passed?
Substituting either $\mathcal{O}_\text{inst}$ or $\mathcal{O}_\text{rub}$ into Equations~\ref{eq:hspp-ratio}--\ref{eq:hspp-ratio-family} yields the instance-level or rubric-level HSPP Ratio, respectively.

\subsection{Measuring performance}\label{subsec:measuring-performance}

Accuracy measures the rate of agreement between the judge's verdicts and the reference across a benchmark $\mathcal{D}$.
As with SPB, this can be instantiated at different granularities depending on the evaluation paradigm.

\paragraph{Instance-level accuracy.}
For all four paradigms (SR, AR, DA, PWC), we measure accuracy through \emph{pairwise concordance}: for each judge, instance, and pair of generators $(\mathcal{G}, \mathcal{G}')$, we check whether the judge's ordering agrees with the reference.
Using the same notation as in Equation~\ref{eq:overest-instance}, a comparison is concordant if $w_{\mathcal{J}}(\mathcal{G}, \mathcal{G}', x) = w^*(\mathcal{G}, \mathcal{G}', x)$, including ties (\textit{i.e.,} both signs are zero).
The \emph{Mean Instance Pairwise Accuracy} (MIPA) for judge $\mathcal{J}$ is:
\begin{equation}\label{eq:mipa}
    \text{MIPA}(\mathcal{J}) = \frac{1}{\binom{|\mathcal{G}|}{2} |\mathcal{D}|} \sum_{\mathcal{G} < \mathcal{G}'} \sum_{x \in \mathcal{D}} \mathbb{1}\{w_{\mathcal{J}}(\mathcal{G}, \mathcal{G}', x) = w^*(\mathcal{G}, \mathcal{G}', x)\}\,,
\end{equation}
where the outer sum runs over all unordered pairs of generators.
For PWC, we run each comparison twice, swapping the presentation order of the two outputs to mitigate position bias~\citep{zheng2023judging}. If one run produces a winner and the other is a tie, the non-tie result stands. 
If both runs are ties or the two runs disagree on the winner, the result is a tie.

\paragraph{Rubric-level accuracy.}
For SR and AR, the \emph{Mean Rubric Accuracy} (MRA) for judge $\mathcal{J}$ is simply the fraction of individual rubric verdicts that match the reference:
\begin{equation}\label{eq:mra}
    \text{MRA}(\mathcal{J}) = \frac{\displaystyle\sum_{\mathcal{G}} \sum_{x \in \mathcal{D}} \sum_{k} \mathbb{1}\big\{b_\mathcal{J}(\mathcal{G}, x, k) = b^*(\mathcal{G}, x, k)\big\}}{\displaystyle\sum_{\mathcal{G}} \sum_{x \in \mathcal{D}} \sum_{k} 1}\,.
\end{equation}

\subsection{Statistical significance}\label{subsec:statistical-significance}

We report 95\% confidence intervals for all metrics, computed via percentile bootstrap with 1000 resamples and shown in gray next to each point estimate.
The resampling unit matches the granularity of the metric: for instance-level metrics, we resample benchmark instances; for rubric-level metrics, we pool the eligible rubric verdicts of all participating generators and resample from this global pool, recomputing the full metric, numerator and denominator alike, on each resample.
When comparing paired quantities (\textit{e.g.,} a committee member against the committee on the member's own outputs, Appendix~\ref{app:ensembles}), we share resamples across the two sides so that the interval on their difference is valid.

\subsection{Models}\label{subsec:models}

We evaluate a broad set of frontier open- and closed-weights models spanning different families and sizes as judges and generators: Gemma 3 27B, 12B, and 4B~\citep{kamath2025gemma}, Llama 4 Maverick and Scout~\citep{meta2025llama4}, Qwen 3 235B, 30B, and 4B~\citep{yang2025qwen3}, GPT-5 and GPT-oss-120B~\citep{agarwal2025gpt}, and Claude 4.5 Sonnet and Haiku.\footnote{We use proprietary models with default reasoning effort, and we do not include them in the AR, DA, and PWC experiments for cost purposes. In Appendix~\ref{app:reasoning}, we find that reasoning effort does not reduce self-preference on IFEval.}

\subsection{Datasets}\label{subsec:datasets}

The metrics in Section~\ref{subsec:measuring-spb} require a ground-truth reference, which is typically unavailable; this is precisely why LLM-as-a-judge is used in the first place.
Thus, inspired by \citet{chen2025llm}, we first conduct our analysis on two benchmarks: 1) IFEval~\citep{zhou2023instruction}, a benchmark of 541 prompts for evaluating instruction-following, where each prompt is paired with one or more programmatically verifiable instructions (\textit{e.g.,} ``no commas''); and 2) LiveCodeBench~\citep{jain2025livecodebench}, a code-generation benchmark.\footnote{We use the data split after 2025-01-04 to limit data contamination, and consider only tests that every model managed to judge, yielding 177 problems and 5,685 distinct tests/rubrics.}
We convert each instruction and each private unit test into binary rubrics, respectively.
This allows us to compute overestimation metrics precisely and to isolate the effect of SPB from confounders like rubric subjectivity (Section~\ref{sec:results-objective}).

To analyze SPB in a subjective setting, we turn to HealthBench~\citep{arora2025healthbench} in Section~\ref{sec:results-subjective}, a benchmark for medical chat assistant evaluation containing 5,000 instances and 48,562 unique binary rubrics.
Rubrics range from accuracy and instruction-following to highly subjective areas like communication tone, and are each associated with a weight toward the final score that the judge does not have access to.
Some rubrics are negative, meaning that the generator is penalized for satisfying them.
Since ground-truth verdicts are unavailable, we use as reference a majority vote among the largest judge of each family (Gemma 3 27B, Llama 4 Maverick, Qwen 3 235B, GPT-oss-120B, and Claude Sonnet 4.5).
All 5 models score higher than 0.6 on HealthBench's own meta-evaluation set (Appendix~\ref{sec:appendix-healthbench-meta-eval}), which is within the range of solid judge performance reported by the authors.
While imperfect, this is nonetheless informative: ensembles of judges have been shown to be more accurate and robust to SPB than individual judges~\citep{verga2024replacing}, which we observe on IFEval (\S\ref{sec:results-objective}).\footnote{Using an ensemble as a reference actually leads to an underestimation of SPB, considering that all model families are represented in the reference (see Appendix~\ref{app:reference-ablations}).}
Appendix~\ref{sec:appendix-prompts} contains details on the prompts used for all evaluation modes and benchmarks.

\section{Self-preference bias with objective rubrics}\label{sec:results-objective}

\begin{table*}[t]
\centering
\setlength{\tabcolsep}{3pt}
\renewcommand{\arraystretch}{1.25}
\footnotesize
\begin{tabular}{l cc cc cc}
\toprule
 & \multicolumn{2}{c}{IFEval} & \multicolumn{2}{c}{LiveCodeBench} & \multicolumn{2}{c}{HealthBench} \\
\cmidrule(lr){2-3} \cmidrule(lr){4-5} \cmidrule(lr){6-7}
Judge & MRA & \makecell{HSPP-Rub.\\(Self)} & MRA & \makecell{HSPP-Rub.\\(Self)} & MRA & \makecell{HSPP-Rub.\\(Self)} \\
\midrule
\textcolor{SwordOrange}{\large{$\boldsymbol{\cdot}$}}Gemma-4B & 0.89\spbci{0.89}{0.90} & 1.03\spbci{1.00}{1.06} & 0.46\spbci{0.45}{0.46} & 1.54\spbci{1.38}{1.73} & 0.71\spbci{0.70}{0.71} & 1.16\spbci{1.14}{1.17} \\
\textcolor{SwordOrange}{\large{$\boldsymbol{\cdot}$}}Gemma-12B & 0.89\spbci{0.88}{0.89} & 1.03\spbci{0.95}{1.11} & 0.74\spbci{0.74}{0.75} & 0.97\spbci{0.93}{1.01} & 0.77\spbci{0.77}{0.77} & 1.03\spbci{1.01}{1.05} \\
\textcolor{SwordOrange}{\large{$\boldsymbol{\cdot}$}}Gemma-27B & 0.89\spbci{0.89}{0.90} & 1.08\spbci{1.02}{1.14} & 0.77\spbci{0.77}{0.78} & 1.03\spbci{0.97}{1.08} & 0.83\spbci{0.83}{0.83} & 1.10\spbci{1.08}{1.12} \\
\addlinespace[2pt]
\textcolor{SwordYellow}{\large{$\boldsymbol{\cdot}$}}Llama-Scout & 0.84\spbci{0.83}{0.85} & 1.05\spbci{0.91}{1.20} & 0.77\spbci{0.77}{0.77} & 1.17\spbci{1.14}{1.20} & 0.83\spbci{0.83}{0.83} & 0.80\spbci{0.79}{0.82} \\
\textcolor{SwordYellow}{\large{$\boldsymbol{\cdot}$}}Llama-Mav & 0.88\spbci{0.87}{0.88} & 1.12\spbci{0.98}{1.27} & 0.79\spbci{0.78}{0.79} & 1.34\spbci{1.30}{1.38} & 0.90\spbci{0.90}{0.90} & 0.71\spbci{0.69}{0.73} \\
\addlinespace[2pt]
\textcolor{SwordBlue}{\large{$\boldsymbol{\cdot}$}}Qwen-4B & 0.85\spbci{0.84}{0.86} & 1.06\spbci{0.92}{1.20} & 0.67\spbci{0.66}{0.67} & 2.21\spbci{2.03}{2.42} & 0.77\spbci{0.77}{0.77} & 0.96\spbci{0.95}{0.98} \\
\textcolor{SwordBlue}{\large{$\boldsymbol{\cdot}$}}Qwen-30B & 0.88\spbci{0.88}{0.89} & 1.05\spbci{0.93}{1.15} & 0.79\spbci{0.78}{0.79} & 1.08\spbci{0.99}{1.18} & 0.81\spbci{0.81}{0.81} & 1.06\spbci{1.04}{1.08} \\
\textcolor{SwordBlue}{\large{$\boldsymbol{\cdot}$}}Qwen-235B & 0.89\spbci{0.88}{0.89} & 1.30\spbci{1.11}{1.48} & 0.65\spbci{0.65}{0.66} & 0.83\spbci{0.62}{1.07} & 0.94\spbci{0.94}{0.94} & 1.12\spbci{1.07}{1.18} \\
\addlinespace[2pt]
\textcolor{SwordSquash}{\large{$\boldsymbol{\cdot}$}}GPT-120B & 0.91\spbci{0.90}{0.91} & 1.30\spbci{0.81}{1.85} & 0.91\spbci{0.91}{0.92} & 5.97\spbci{4.91}{7.18} & 0.87\spbci{0.87}{0.87} & 1.00\spbci{0.94}{1.06} \\
\textcolor{SwordSquash}{\large{$\boldsymbol{\cdot}$}}GPT-5 & 0.91\spbci{0.91}{0.92} & 1.47\spbci{0.48}{2.53} & 0.97\spbci{0.97}{0.97} & 20.15\spbci{16.57}{24.58} & 0.86\spbci{0.86}{0.86} & 1.54\spbci{1.44}{1.64} \\
\addlinespace[2pt]
\textcolor{SwordPink}{\large{$\boldsymbol{\cdot}$}}Claude-Haiku & 0.81\spbci{0.80}{0.82} & 1.15\spbci{0.88}{1.44} & 0.92\spbci{0.92}{0.93} & 1.71\spbci{1.57}{1.85} & 0.84\spbci{0.84}{0.84} & 0.90\spbci{0.86}{0.93} \\
\textcolor{SwordPink}{\large{$\boldsymbol{\cdot}$}}Claude-Sonnet & 0.84\spbci{0.83}{0.85} & 0.98\spbci{0.64}{1.32} & 0.95\spbci{0.95}{0.95} & 1.78\spbci{1.58}{1.99} & 0.88\spbci{0.88}{0.88} & 0.93\spbci{0.86}{1.01} \\
\bottomrule
\end{tabular}

\caption{Rubric-level metrics per judge across IFEval, LiveCodeBench, and HealthBench (SR), with 95\% confidence intervals (rubric-level bootstrap, 1000 resamples) in grey. MRA = Mean Rubric Accuracy; HSPP-Rub.\ (Self) = rubric-level HSPP ratio for self-generated outputs. Family-level ratios are in Appendix~\ref{app:family}.}
\label{tab:rubric-metrics}
\end{table*}

\begin{figure*}[t]
    \centering
    \begin{minipage}[t]{0.49\textwidth}\centering
    \begin{tikzpicture}
    \begin{axis}[
        spb axis,
        width=0.95\textwidth,
        height=5.5cm,
        xlabel={Mean Instance Pairwise Accuracy},
        ylabel={HSPP-Instance ratio},
        xmin=0.3516, xmax=0.9666,
        ymin=0.7, ymax=1.7,
        xtick={0.3000, 0.4000, 0.5000, 0.6000, 0.7000, 0.8000, 0.9000, 1.0000},
        ytick={0.7, 0.8, 0.9, 1.0, 1.1, 1.2, 1.3, 1.4, 1.5, 1.6, 1.7},
    ]

    \addplot[gray!50, dashed, thin, forget plot] coordinates {(0.3516,1.0) (0.9666,1.0)};

    \addplot[only marks, mark=square*, mark size=2.75pt, SwordOrange, fill opacity=1.0, draw=none] coordinates {(0.651515,1.098419) (0.409091,1.034777) (0.469697,1.033423) (0.803030,1.197693) (0.621212,1.012749) (0.863636,1.168879) (0.772727,1.048560) (0.742424,1.018502) (0.893939,1.092118)};
    \addplot[only marks, mark=square*, mark size=2.75pt, SwordOrange, fill opacity=0.35, draw=none, forget plot] coordinates {(0.954545,1.657579) (0.833333,1.091154) (0.893939,1.160305)};

    \addplot[only marks, mark=triangle*, mark size=3.3pt, SwordYellow, fill opacity=1.0, draw=none] coordinates {(0.712121,1.079129) (0.575758,1.053495) (0.363636,1.006497) (0.742424,1.206061) (0.560606,1.034411) (0.878788,1.080019) (0.681818,1.040730) (0.681818,1.016859) (0.909091,1.020123)};

    \addplot[only marks, mark=diamond*, mark size=3.3pt, SwordBlue, fill opacity=1.0, draw=none] coordinates {(0.560606,1.072052) (0.530303,1.027115) (0.393939,1.013932) (0.742424,1.164290) (0.560606,1.065368) (0.818182,1.067895) (0.681818,1.000950) (0.696970,1.022464) (0.909091,0.954023)};

    \addplot[only marks, mark=*, mark size=2.75pt, SwordSquash, fill opacity=1.0, draw=none] coordinates {(0.515152,1.568518) (0.424242,1.364562) (0.378788,1.326796) (0.590909,0.819855) (0.575758,0.704954) (0.666667,1.653342) (0.636364,1.453501) (0.636364,1.304302) (0.833333,0.962777)};

    \end{axis}
    \end{tikzpicture}
    \end{minipage}%
    \hfill%
    \begin{minipage}[t]{0.49\textwidth}\centering
    \begin{tikzpicture}
    \begin{axis}[
        spb axis,
        width=0.95\textwidth,
        height=5.5cm,
        xlabel={Mean Rubric Accuracy},
        ylabel={HSPP-Rubric ratio},
        xmin=0.8079, xmax=0.9150,
        ymin=0.7, ymax=1.7,
        xtick={0.8000, 0.8200, 0.8400, 0.8600, 0.8800, 0.9000, 0.9200},
        ytick={0.7, 0.8, 0.9, 1.0, 1.1, 1.2, 1.3, 1.4, 1.5, 1.6, 1.7},
    ]

    \addplot[gray!50, dashed, thin, forget plot] coordinates {(0.8079,1.0) (0.9150,1.0)};

    \node[font=\scriptsize, anchor=north west, align=left] at (rel axis cs:0.02,0.97) {$\blacktriangle$ times more likely to overestimate\\self than others};

    \addplot[only marks, mark=square*, mark size=2.75pt, SwordOrange, fill opacity=1.0, draw=none] coordinates {(0.894484,1.083803) (0.887090,1.029708) (0.893985,1.029534) (0.876499,1.121061) (0.841827,1.047547) (0.887290,1.297454) (0.881994,1.045161) (0.849720,1.060319) (0.905276,1.304868)};
    \addplot[only marks, mark=square*, mark size=2.75pt, SwordOrange, fill opacity=0.35, draw=none, forget plot] coordinates {(0.912770,1.465707) (0.810152,1.150998) (0.840328,0.981923)};

    \addplot[only marks, mark=triangle*, mark size=3.3pt, SwordYellow, fill opacity=1.0, draw=none] coordinates {(0.898082,1.066819) (0.897682,1.062263) (0.895983,1.011530) (0.877998,1.149975) (0.858813,1.046642) (0.890388,1.176885) (0.898581,1.066208) (0.868106,1.069976) (0.906175,1.068651)};

    \end{axis}
    \end{tikzpicture}
    \end{minipage}%

    \vspace{0.3em}

    \begin{tikzpicture}
    \begin{axis}[
        hide axis,
        width=\textwidth,
        height=1.4cm,
        scale only axis,
        xmin=0, xmax=1, ymin=0, ymax=1,
        legend style={
            at={(0.5,0.5)},
            anchor=center,
            legend columns=2,
            column sep=0.3cm,
            draw=none,
            fill=none,
        },
        legend image post style={mark size=3pt},
    ]
    \addlegendimage{only marks, mark=square*, mark size=2.75pt, SwordOrange, fill opacity=1.0, draw=none}
    \addlegendentry{Single Rubric (SR)}
    \addlegendimage{only marks, mark=triangle*, mark size=3.3pt, SwordYellow, fill opacity=1.0, draw=none}
    \addlegendentry{All Rubrics (AR)}
    \addlegendimage{only marks, mark=diamond*, mark size=3.3pt, SwordBlue, fill opacity=1.0, draw=none}
    \addlegendentry{Direct Assessment (DA)}
    \addlegendimage{only marks, mark=*, mark size=2.75pt, SwordSquash, fill opacity=1.0, draw=none}
    \addlegendentry{Pairwise Comparison (PWC)}
    \end{axis}
    \end{tikzpicture}
    \caption{Performance vs.\ self-preference bias on IFEval across judge paradigms. SR shows also shows GPT-5 and Claude models (light orange squares), which we did not run in other methods for cost purposes. \textit{Left:} Instance-level (MIPA vs.\ HSPP-Instance ratio). \textit{Right:} Rubric-level (MRA vs.\ HSPP-Rubric ratio).}
    \label{fig:scatter-ifeval-combined}
\end{figure*}

\begin{figure*}[t]
    \centering
    \begin{minipage}[t]{0.49\textwidth}\centering
    \begin{tikzpicture}
    \begin{axis}[
        spb axis,
        width=0.85\textwidth,
        height=4.5cm,
        ylabel={HSPP-Rubric ratio},
        xmin=-0.3, xmax=1.3,
        ymin=0.9, ymax=1.55,
        xtick={0, 1},
        xticklabels={Individual, Committee},
        ytick={0.90, 1.00, 1.10, 1.20, 1.30, 1.40, 1.50},
        x tick label style={font=\small},
        clip=false,
    ]

    \addplot[gray!50, dashed, line width=0.6pt, forget plot] coordinates {(-0.3,1.0) (1.3,1.0)};

    \addplot[SwordOrange, thick, no markers, forget plot, shorten <=2.5pt] coordinates {(0,1.083803) (1,1.081340)};
    \addplot[only marks, mark=o, mark size=2.5pt, draw=SwordOrange, line width=0.9pt, forget plot] coordinates {(0,1.083803)};
    \addplot[only marks, mark=*, mark size=2.5pt, SwordOrange, draw=none, forget plot] coordinates {(1,1.081340)};
    \addplot[SwordYellow, thick, no markers, forget plot, shorten <=2.5pt] coordinates {(0,1.121061) (1,0.968324)};
    \addplot[only marks, mark=o, mark size=2.5pt, draw=SwordYellow, line width=0.9pt, forget plot] coordinates {(0,1.121061)};
    \addplot[only marks, mark=*, mark size=2.5pt, SwordYellow, draw=none, forget plot] coordinates {(1,0.968324)};
    \addplot[SwordBlue, thick, no markers, forget plot, shorten <=2.5pt] coordinates {(0,1.297454) (1,1.193561)};
    \addplot[only marks, mark=o, mark size=2.5pt, draw=SwordBlue, line width=0.9pt, forget plot] coordinates {(0,1.297454)};
    \addplot[only marks, mark=*, mark size=2.5pt, SwordBlue, draw=none, forget plot] coordinates {(1,1.193561)};
    \addplot[SwordSquash, thick, no markers, forget plot, shorten <=2.5pt] coordinates {(0,1.465707) (1,1.095729)};
    \addplot[only marks, mark=o, mark size=2.5pt, draw=SwordSquash, line width=0.9pt, forget plot] coordinates {(0,1.465707)};
    \addplot[only marks, mark=*, mark size=2.5pt, SwordSquash, draw=none, forget plot] coordinates {(1,1.095729)};
    \addplot[SwordPink, thick, no markers, forget plot, shorten <=2.5pt] coordinates {(0,0.981923) (1,0.988759)};
    \addplot[only marks, mark=o, mark size=2.5pt, draw=SwordPink, line width=0.9pt, forget plot] coordinates {(0,0.981923)};
    \addplot[only marks, mark=*, mark size=2.5pt, SwordPink, draw=none, forget plot] coordinates {(1,0.988759)};

    \node[font=\scriptsize, anchor=west, SwordBlue!80!black] at (axis cs:1.05,1.217608) {Qwen-235B};
    \node[font=\scriptsize, anchor=west, SwordSquash!80!black] at (axis cs:1.05,1.119777) {GPT-5};
    \node[font=\scriptsize, anchor=west, SwordOrange!80!black] at (axis cs:1.05,1.061277) {Gemma-27B};
    \node[font=\scriptsize, anchor=west, SwordPink!80!black] at (axis cs:1.05,1.002777) {Claude-Sonnet};
    \node[font=\scriptsize, anchor=west, SwordYellow!80!black] at (axis cs:1.05,0.944277) {Llama-Maverick};

    \end{axis}
    \end{tikzpicture}
    \end{minipage}%
    \hfill%
    \begin{minipage}[t]{0.49\textwidth}\centering
    \begin{tikzpicture}
    \begin{axis}[
        spb axis,
        width=0.85\textwidth,
        height=4.5cm,
        ylabel={Mean Rubric Accuracy},
        xmin=-0.3, xmax=1.3,
        ymin=0.83, ymax=0.93,
        xtick={0, 1},
        xticklabels={Individual, Committee},
        ytick={0.83, 0.85, 0.87, 0.89, 0.91, 0.93},
        x tick label style={font=\small},
        clip=false,
    ]

    \addplot[SwordOrange, thick, no markers, forget plot, shorten <=2.5pt] coordinates {(0,0.894484) (1,0.911671)};
    \addplot[only marks, mark=o, mark size=2.5pt, draw=SwordOrange, line width=0.9pt, forget plot] coordinates {(0,0.894484)};
    \addplot[only marks, mark=*, mark size=2.5pt, SwordOrange, draw=none, forget plot] coordinates {(1,0.911671)};
    \addplot[SwordYellow, thick, no markers, forget plot, shorten <=2.5pt] coordinates {(0,0.876499) (1,0.911671)};
    \addplot[only marks, mark=o, mark size=2.5pt, draw=SwordYellow, line width=0.9pt, forget plot] coordinates {(0,0.876499)};
    \addplot[only marks, mark=*, mark size=2.5pt, SwordYellow, draw=none, forget plot] coordinates {(1,0.911671)};
    \addplot[SwordBlue, thick, no markers, forget plot, shorten <=2.5pt] coordinates {(0,0.887290) (1,0.911671)};
    \addplot[only marks, mark=o, mark size=2.5pt, draw=SwordBlue, line width=0.9pt, forget plot] coordinates {(0,0.887290)};
    \addplot[only marks, mark=*, mark size=2.5pt, SwordBlue, draw=none, forget plot] coordinates {(1,0.911671)};
    \addplot[SwordSquash, thick, no markers, forget plot, shorten <=2.5pt] coordinates {(0,0.912770) (1,0.911671)};
    \addplot[only marks, mark=o, mark size=2.5pt, draw=SwordSquash, line width=0.9pt, forget plot] coordinates {(0,0.912770)};
    \addplot[only marks, mark=*, mark size=2.5pt, SwordSquash, draw=none, forget plot] coordinates {(1,0.911671)};
    \addplot[SwordPink, thick, no markers, forget plot, shorten <=2.5pt] coordinates {(0,0.840328) (1,0.911671)};
    \addplot[only marks, mark=o, mark size=2.5pt, draw=SwordPink, line width=0.9pt, forget plot] coordinates {(0,0.840328)};
    \addplot[only marks, mark=*, mark size=2.5pt, SwordPink, draw=none, forget plot] coordinates {(1,0.911671)};

    \node[font=\scriptsize, anchor=west, SwordSquash!80!black] at (axis cs:1.05,0.929671) {GPT-5};
    \node[font=\scriptsize, anchor=west, SwordOrange!80!black] at (axis cs:1.05,0.920671) {Gemma-27B};
    \node[font=\scriptsize, anchor=west, SwordBlue!80!black] at (axis cs:1.05,0.911671) {Qwen-235B};
    \node[font=\scriptsize, anchor=west, SwordYellow!80!black] at (axis cs:1.05,0.902671) {Llama-Maverick};
    \node[font=\scriptsize, anchor=west, SwordPink!80!black] at (axis cs:1.05,0.893671) {Claude-Sonnet};

    \end{axis}
    \end{tikzpicture}
    \end{minipage}%

    \vspace{0.3em}

    \begin{tikzpicture}
    \begin{axis}[
        hide axis,
        width=\textwidth,
        height=1.4cm,
        scale only axis,
        xmin=0, xmax=1, ymin=0, ymax=1,
        legend style={
            at={(0.5,0.5)},
            anchor=center,
            legend columns=3,
            column sep=0.3cm,
            draw=none,
            fill=none,
        },
        legend image post style={mark size=2.5pt},
    ]
    \addlegendimage{SwordOrange, thick, mark=*, mark options={fill=SwordOrange, draw=none}}
    \addlegendentry{Gemma-27B}
    \addlegendimage{SwordYellow, thick, mark=*, mark options={fill=SwordYellow, draw=none}}
    \addlegendentry{Llama-Maverick}
    \addlegendimage{SwordBlue, thick, mark=*, mark options={fill=SwordBlue, draw=none}}
    \addlegendentry{Qwen-235B}
    \addlegendimage{SwordSquash, thick, mark=*, mark options={fill=SwordSquash, draw=none}}
    \addlegendentry{GPT-5}
    \addlegendimage{SwordPink, thick, mark=*, mark options={fill=SwordPink, draw=none}}
    \addlegendentry{Claude-Sonnet}
    \end{axis}
    \end{tikzpicture}
    \caption{Impact of a 5-family committee on rubric-level self-preference bias (left) and accuracy (right) on IFEval (SR). Lines connect each committee member's individual performance to its committee-aggregated performance.}
    \label{fig:slope-ifeval-committee}
\end{figure*}

\begin{figure*}[t]
    \centering
    \begin{minipage}[t]{0.49\textwidth}\centering
    \begin{tikzpicture}
    \begin{axis}[
        spb axis,
        width=\textwidth,
        height=4.5cm,
        xlabel={Agreement threshold},
        ylabel={HSPP-Instance ratio},
        ymin=0.800000, ymax=1.600000,
        xmin=0.49, xmax=1.01,
        xtick={0.5, 0.6, 0.7, 0.8, 0.9, 1.0},
        xticklabels={0.5, 0.6, 0.7, 0.8, 0.9, 1.0},
        ytick={0.8000, 0.9000, 1.0000, 1.1000, 1.2000, 1.3000, 1.4000, 1.5000, 1.6000},
    ]

    \addplot[gray!50, dashed, thin, forget plot] coordinates {(0.49,1.0) (1.01,1.0)};

    \addplot[SwordOrange, line width=1.2pt] coordinates {(0.50,1.465081) (0.51,1.480249) (0.52,1.458994) (0.53,1.486147) (0.54,1.540346) (0.55,1.553403) (0.56,1.499637) (0.57,1.437465) (0.58,1.437445) (0.59,1.439718) (0.60,1.286777) (0.61,1.284955) (0.62,1.288277) (0.63,1.278616) (0.64,1.282091) (0.65,1.284096) (0.66,1.291993) (0.67,1.291030) (0.68,1.279689) (0.69,1.278304) (0.70,1.262271) (0.71,1.294319) (0.72,1.281174) (0.73,1.288219) (0.74,1.303679) (0.75,1.291095) (0.76,1.418961) (0.77,1.221296) (0.78,1.193834) (0.79,1.204416) (0.80,1.209119) (0.81,1.185206) (0.82,1.143791) (0.83,1.140415) (0.84,1.117823) (0.85,1.075547) (0.86,1.063821) (0.87,1.023307) (0.88,0.972506) (0.89,0.932665) (0.90,0.897319) (0.91,0.965870) (0.92,0.943906) (0.93,1.035555) (0.94,0.958858) (0.95,1.032376) (0.96,1.016499) (0.97,1.110827) (0.98,1.074246)};

    \end{axis}
    \end{tikzpicture}
    \end{minipage}%
    \hfill%
    \begin{minipage}[t]{0.49\textwidth}\centering
    \begin{tikzpicture}
    \begin{axis}[
        spb axis,
        width=\textwidth,
        height=4.5cm,
        xlabel={Agreement threshold},
        ylabel={HSPP-Rubric ratio},
        ymin=0.800000, ymax=1.600000,
        xmin=0.49, xmax=1.01,
        xtick={0.5, 0.6, 0.7, 0.8, 0.9, 1.0},
        xticklabels={0.5, 0.6, 0.7, 0.8, 0.9, 1.0},
        ytick={0.8000, 0.9000, 1.0000, 1.1000, 1.2000, 1.3000, 1.4000, 1.5000, 1.6000},
    ]

    \addplot[gray!50, dashed, thin, forget plot] coordinates {(0.49,1.0) (1.01,1.0)};

    \addplot[SwordOrange, line width=1.2pt] coordinates {(0.50,1.136532) (0.51,1.134646) (0.52,1.128316) (0.53,1.135393) (0.54,1.140296) (0.55,1.146581) (0.56,1.150886) (0.57,1.156142) (0.58,1.154634) (0.59,1.144276) (0.60,1.137822) (0.61,1.131210) (0.62,1.134520) (0.63,1.136726) (0.64,1.144258) (0.65,1.133508) (0.66,1.134302) (0.67,1.130997) (0.68,1.125589) (0.69,1.133127) (0.70,1.124010) (0.71,1.133498) (0.72,1.138361) (0.73,1.138070) (0.74,1.155973) (0.75,1.154340) (0.76,1.150600) (0.77,1.139219) (0.78,1.160326) (0.79,1.155194) (0.80,1.165960) (0.81,1.164532) (0.82,1.151463) (0.83,1.132095) (0.84,1.093139) (0.85,1.083511) (0.86,1.086730) (0.87,1.081727) (0.88,1.083328) (0.89,1.118677) (0.90,1.090035) (0.91,1.115412) (0.92,1.104549) (0.93,1.058563) (0.94,1.038850) (0.95,1.035578) (0.96,1.037074) (0.97,1.049973) (0.98,1.024772)};

    \node[font=\scriptsize, anchor=north west, align=left] at (rel axis cs:0.02,0.97) {$\blacktriangle$ times more likely to overestimate\\self than others};

    \end{axis}
    \end{tikzpicture}
    \end{minipage}%
    \caption{HSPP-Instance (left) and HSPP-Rubric (right) when filtering rubrics by inter-judge pairwise agreement threshold on IFEval (SR, 12 judges, mean over all judges).}
    \label{fig:lineplot-ifeval-agreement-threshold}
\end{figure*}

\paragraph{Judges favor themselves and models from their family even with objective rubrics.}

The LiveCodeBench part of Table~\ref{tab:rubric-metrics} shows that most judges have a rubric-level HSPP ratio significantly greater than 1.
In the worst case (GPT-5), the judge can be 20 times more likely to overestimate its own outputs than those of unrelated models.
On IFEval, all but one judge exhibits a mean HSPP ratio above 1, though statistical significance only holds for two systems.
Table \ref{tab:app-family} shows similar results at the family level.
This finding suggests that perfect evaluation criteria alone are not sufficient to prevent SPB at individual or family levels.
Elimination of SPB may instead require more intricate strategies (\textit{e.g.,} model training, activation steering~\citep{roytburg2025breaking}), rather than just careful evaluation design.

\paragraph{Rubric-based evaluation is more robust than pairwise comparison, but not necessarily direct assessment.}

Figure~\ref{fig:scatter-ifeval-combined} shows that, while SPB is present across all paradigms, it is significantly more severe in PWC than in RB and DA.
This could be because PWC directly pits the judge's output against that of an opponent, making it easier for the judge to identify and favor its own output.
Indeed, \citet{panickssery2024llm} have shown that SPB is related to the judge's ability to identify its own outputs, which we also find in Appendix~\ref{app:mechanisms}.
At the rubric level, presenting one rubric at a time elicits slightly higher SPB than presenting all at once, possibly because the narrow scope makes it easier for the judge to rationalize away failures on its own outputs.

\paragraph{Ensembling judgments helps mitigate and predict self-preference.}

In Figure~\ref{fig:slope-ifeval-committee}, we show instance- and rubric-level accuracy and SPB metrics on IFEval for the largest judge of each family, as well as for a majority-vote ensemble of all five judges.
A judge's committee-aggregated performance and SPB are computed in the same way as for individual judges, but using the ensemble's majority vote as the prediction instead of the individual verdict.
Across the board, ensembling leads to the same or higher accuracy and lower SPB, though it does not fully eliminate SPB.
This is in line with the findings of~\citet{verga2024replacing} and suggests that ensembling is a practical strategy for mitigating SPB, though not sufficient on its own.
In Appendix~\ref{app:ensembles}, we show that the main drivers of ensemble-based mitigation are the inclusion of high-quality judges, and, to a lesser extent, family diversity.
Furthermore, we find that removing rubrics with low inter-judge agreement can be an effective way to reduce SPB, as shown in Figure~\ref{fig:lineplot-ifeval-agreement-threshold}, indicating that agreement can be a useful signal for flagging problematic rubrics.

\begin{table}[t]
\centering
\setlength{\tabcolsep}{3pt}
\renewcommand{\arraystretch}{1.25}
\footnotesize
\begin{tabular}{l c cccc}
\toprule
 & & \multicolumn{4}{c}{Mean-MRA bin (proxy for evaluation ``hardness'')} \\
\cmidrule(lr){3-6}
Judge & Original & 0.5--0.6 & 0.6--0.7 & 0.7--0.8 & 0.8--0.9 \\
\midrule
\textcolor{SwordSquash}{\large{$\boldsymbol{\cdot}$}}GPT-5 & 20.15\spbci{16.57}{24.58} & 13.74\spbci{9.94}{19.29} & 14.66\spbci{8.94}{21.48} & 7.36\spbci{1.50}{15.55} & 0.00\spbci{0.00}{0.00} \\
\textcolor{SwordSquash}{\large{$\boldsymbol{\cdot}$}}GPT-120B & 5.97\spbci{4.91}{7.18} & 1.13\spbci{0.59}{1.84} & 1.98\spbci{0.83}{3.24} & 3.98\spbci{1.48}{7.33} & 7.35\spbci{1.34}{21.10} \\
\textcolor{SwordPink}{\large{$\boldsymbol{\cdot}$}}Claude-Sonnet & 1.78\spbci{1.58}{1.99} & 1.86\spbci{1.56}{2.14} & 1.33\spbci{1.11}{1.60} & 2.06\spbci{1.45}{2.89} & 2.76\spbci{0.89}{5.22} \\
\textcolor{SwordBlue}{\large{$\boldsymbol{\cdot}$}}Qwen-235B & 0.83\spbci{0.62}{1.07} & 0.78\spbci{0.40}{1.23} & 0.93\spbci{0.58}{1.38} & 0.25\spbci{0.06}{0.61} & 2.28\spbci{0.79}{4.49} \\
\bottomrule
\end{tabular}

\caption{LiveCodeBench rubric-level HSPP (Self) stratified by generation hardness: ratios are recomputed using only (generator, instance) cells whose mean rubric accuracy across all 12 judges falls in the bin. ``Original'' is the unstratified ratio from Table~\ref{tab:rubric-metrics}. 
Null values occur when there are no instances in the bin for that generator.}
\label{tab:mra-strata-lcb}
\end{table}

\paragraph{Self-preference resists generator capability confounds.} One possible explanation for the high HSPP values of models like GPT-5 on LiveCodeBench is that they are simply better at the task, so their mistakes are harder to judge and, therefore, more likely to be overestimated.
To control for this, we stratify instances and their respective completion by mean rubric accuracy across all 12 judges, which serves as a proxy for their ``hardness''.
Concretely, instances where judges correctly classify more rubrics are considered easier, while those that judges incorrectly classify more rubrics wrong are considered harder.
We expect better generators to have a higher fraction of failed unit tests in the harder bins, and worse generators to have more failed tests in the easier bins; this is indeed the case, as we show in Appendix Table~\ref{tab:app-fail-hardness-lcb}.
Thus, in Table~\ref{tab:mra-strata-lcb}, we recompute the rubric-level HSPP ratio for the 4 best-performing models on LiveCodeBench. 
If self-preference were an artifact of hard-to-judge generations, the ratio should collapse to 1 within each hardness bin. 
Instead, we observe that, with the exception of Qwen 235B, models maintain a high HSPP ratio across bins, indicating that self-preference persists beyond generator capability.

\section{Self-preference bias with subjective rubrics}\label{sec:results-subjective}

\begin{figure*}[t]
    \centering
    \begin{tikzpicture}[
        cell/.style={minimum width=0.82cm, minimum height=0.36cm, rounded corners=0.12cm, inner sep=0pt, font=\scriptsize},
        self cell/.style={cell, line width=0.6pt},
        header/.style={font=\scriptsize\bfseries, minimum width=0.7cm, minimum height=0.36cm, text depth=0.25ex},
    ]
    \def\colsep{0.95}
    \def\rowsep{0.45}

    \node[header, anchor=east, font=\scriptsize\itshape] at (-0.3, 0.75) {Generator $\rightarrow$};
    \node[header, anchor=east, font=\scriptsize\itshape] at (-0.3, 0.35) {Judge $\downarrow$};
    \node[header, font=\scriptsize, rotate=90, anchor=west] at (0.5*\colsep, 0.35) {Gemma\mbox{-}4B};
    \node[header, font=\scriptsize, rotate=90, anchor=west] at (1.5*\colsep, 0.35) {Gemma\mbox{-}12B};
    \node[header, font=\scriptsize, rotate=90, anchor=west] at (2.5*\colsep, 0.35) {Gemma\mbox{-}27B};
    \node[header, font=\scriptsize, rotate=90, anchor=west] at (3.5*\colsep, 0.35) {Llama\mbox{-}Scout};
    \node[header, font=\scriptsize, rotate=90, anchor=west] at (4.5*\colsep, 0.35) {Llama\mbox{-}Mav};
    \node[header, font=\scriptsize, rotate=90, anchor=west] at (5.5*\colsep, 0.35) {Qwen\mbox{-}4B};
    \node[header, font=\scriptsize, rotate=90, anchor=west] at (6.5*\colsep, 0.35) {Qwen\mbox{-}30B};
    \node[header, font=\scriptsize, rotate=90, anchor=west] at (7.5*\colsep, 0.35) {Qwen\mbox{-}235B};
    \node[header, font=\scriptsize, rotate=90, anchor=west] at (8.5*\colsep, 0.35) {GPT\mbox{-}120B};
    \node[header, font=\scriptsize, rotate=90, anchor=west] at (9.5*\colsep, 0.35) {GPT\mbox{-}5};
    \node[header, font=\scriptsize, rotate=90, anchor=west] at (10.5*\colsep, 0.35) {Claude\mbox{-}Haiku};
    \node[header, font=\scriptsize, rotate=90, anchor=west] at (11.5*\colsep, 0.35) {Claude\mbox{-}Sonnet};

    \node[header, anchor=east, font=\scriptsize] at (-0.15, -0*\rowsep) {Gemma\mbox{-}4B};
    \node[self cell, fill=SwordOrange!70, draw=SwordNoir] at (0.5*\colsep, -0*\rowsep) {\textbf{10.90}};
    \node[cell, fill=SwordOrange!54] at (1.5*\colsep, -0*\rowsep) {8.46};
    \node[cell, fill=SwordOrange!37] at (2.5*\colsep, -0*\rowsep) {5.81};
    \node[cell, fill=SwordOrange!10] at (3.5*\colsep, -0*\rowsep) {1.55};
    \node[cell, fill=SwordOrange!8] at (4.5*\colsep, -0*\rowsep) {0.48};
    \node[cell, fill=SwordOrange!8] at (5.5*\colsep, -0*\rowsep) {0.82};
    \node[cell, fill=SwordBlue!13] at (6.5*\colsep, -0*\rowsep) {-2.07};
    \node[cell, fill=SwordBlue!34] at (7.5*\colsep, -0*\rowsep) {-5.33};
    \node[cell, fill=SwordOrange!25] at (8.5*\colsep, -0*\rowsep) {3.85};
    \node[cell, fill=SwordBlue!80] at (9.5*\colsep, -0*\rowsep) {-12.53};
    \node[cell, fill=SwordBlue!32] at (10.5*\colsep, -0*\rowsep) {-4.99};
    \node[cell, fill=SwordBlue!44] at (11.5*\colsep, -0*\rowsep) {-6.94};
    \node[header, anchor=east, font=\scriptsize] at (-0.15, -1*\rowsep) {Gemma\mbox{-}12B};
    \node[cell, fill=SwordOrange!19] at (0.5*\colsep, -1*\rowsep) {3.05};
    \node[self cell, fill=SwordBlue!13, draw=SwordNoir] at (1.5*\colsep, -1*\rowsep) {\textbf{-2.07}};
    \node[cell, fill=SwordBlue!15] at (2.5*\colsep, -1*\rowsep) {-2.37};
    \node[cell, fill=SwordOrange!38] at (3.5*\colsep, -1*\rowsep) {5.92};
    \node[cell, fill=SwordOrange!35] at (4.5*\colsep, -1*\rowsep) {5.42};
    \node[cell, fill=SwordOrange!12] at (5.5*\colsep, -1*\rowsep) {1.86};
    \node[cell, fill=SwordBlue!8] at (6.5*\colsep, -1*\rowsep) {-0.58};
    \node[cell, fill=SwordBlue!18] at (7.5*\colsep, -1*\rowsep) {-2.77};
    \node[cell, fill=SwordBlue!37] at (8.5*\colsep, -1*\rowsep) {-5.72};
    \node[cell, fill=SwordBlue!52] at (9.5*\colsep, -1*\rowsep) {-8.14};
    \node[cell, fill=SwordOrange!23] at (10.5*\colsep, -1*\rowsep) {3.60};
    \node[cell, fill=SwordOrange!11] at (11.5*\colsep, -1*\rowsep) {1.79};
    \node[header, anchor=east, font=\scriptsize] at (-0.15, -2*\rowsep) {Gemma\mbox{-}27B};
    \node[cell, fill=SwordOrange!28] at (0.5*\colsep, -2*\rowsep) {4.40};
    \node[cell, fill=SwordBlue!8] at (1.5*\colsep, -2*\rowsep) {-0.62};
    \node[self cell, fill=SwordBlue!8, draw=SwordNoir] at (2.5*\colsep, -2*\rowsep) {\textbf{-1.08}};
    \node[cell, fill=SwordOrange!28] at (3.5*\colsep, -2*\rowsep) {4.43};
    \node[cell, fill=SwordOrange!26] at (4.5*\colsep, -2*\rowsep) {4.11};
    \node[cell, fill=SwordOrange!11] at (5.5*\colsep, -2*\rowsep) {1.72};
    \node[cell, fill=SwordBlue!8] at (6.5*\colsep, -2*\rowsep) {-0.56};
    \node[cell, fill=SwordBlue!15] at (7.5*\colsep, -2*\rowsep) {-2.28};
    \node[cell, fill=SwordBlue!33] at (8.5*\colsep, -2*\rowsep) {-5.20};
    \node[cell, fill=SwordBlue!48] at (9.5*\colsep, -2*\rowsep) {-7.52};
    \node[cell, fill=SwordOrange!15] at (10.5*\colsep, -2*\rowsep) {2.31};
    \node[cell, fill=SwordOrange!8] at (11.5*\colsep, -2*\rowsep) {0.27};
    \node[header, anchor=east, font=\scriptsize] at (-0.15, -3*\rowsep) {Llama\mbox{-}Scout};
    \node[cell, fill=SwordOrange!21] at (0.5*\colsep, -3*\rowsep) {3.23};
    \node[cell, fill=SwordBlue!8] at (1.5*\colsep, -3*\rowsep) {-1.06};
    \node[cell, fill=SwordBlue!8] at (2.5*\colsep, -3*\rowsep) {-0.65};
    \node[self cell, fill=SwordOrange!17, draw=SwordNoir] at (3.5*\colsep, -3*\rowsep) {\textbf{2.62}};
    \node[cell, fill=SwordOrange!11] at (4.5*\colsep, -3*\rowsep) {1.70};
    \node[cell, fill=SwordOrange!16] at (5.5*\colsep, -3*\rowsep) {2.51};
    \node[cell, fill=SwordOrange!8] at (6.5*\colsep, -3*\rowsep) {0.71};
    \node[cell, fill=SwordBlue!8] at (7.5*\colsep, -3*\rowsep) {-0.71};
    \node[cell, fill=SwordBlue!23] at (8.5*\colsep, -3*\rowsep) {-3.56};
    \node[cell, fill=SwordBlue!33] at (9.5*\colsep, -3*\rowsep) {-5.22};
    \node[cell, fill=SwordOrange!8] at (10.5*\colsep, -3*\rowsep) {1.10};
    \node[cell, fill=SwordBlue!8] at (11.5*\colsep, -3*\rowsep) {-0.69};
    \node[header, anchor=east, font=\scriptsize] at (-0.15, -4*\rowsep) {Llama\mbox{-}Mav};
    \node[cell, fill=SwordOrange!21] at (0.5*\colsep, -4*\rowsep) {3.27};
    \node[cell, fill=SwordOrange!8] at (1.5*\colsep, -4*\rowsep) {0.66};
    \node[cell, fill=SwordOrange!8] at (2.5*\colsep, -4*\rowsep) {0.01};
    \node[cell, fill=SwordOrange!8] at (3.5*\colsep, -4*\rowsep) {0.41};
    \node[self cell, fill=SwordBlue!8, draw=SwordNoir] at (4.5*\colsep, -4*\rowsep) {\textbf{-0.19}};
    \node[cell, fill=SwordOrange!8] at (5.5*\colsep, -4*\rowsep) {1.03};
    \node[cell, fill=SwordBlue!8] at (6.5*\colsep, -4*\rowsep) {-0.06};
    \node[cell, fill=SwordBlue!9] at (7.5*\colsep, -4*\rowsep) {-1.43};
    \node[cell, fill=SwordBlue!8] at (8.5*\colsep, -4*\rowsep) {-0.91};
    \node[cell, fill=SwordBlue!21] at (9.5*\colsep, -4*\rowsep) {-3.26};
    \node[cell, fill=SwordOrange!8] at (10.5*\colsep, -4*\rowsep) {0.57};
    \node[cell, fill=SwordBlue!8] at (11.5*\colsep, -4*\rowsep) {-0.11};
    \node[header, anchor=east, font=\scriptsize] at (-0.15, -5*\rowsep) {Qwen\mbox{-}4B};
    \node[cell, fill=SwordOrange!26] at (0.5*\colsep, -5*\rowsep) {4.06};
    \node[cell, fill=SwordBlue!12] at (1.5*\colsep, -5*\rowsep) {-1.93};
    \node[cell, fill=SwordBlue!16] at (2.5*\colsep, -5*\rowsep) {-2.49};
    \node[cell, fill=SwordOrange!44] at (3.5*\colsep, -5*\rowsep) {6.93};
    \node[cell, fill=SwordOrange!45] at (4.5*\colsep, -5*\rowsep) {7.12};
    \node[self cell, fill=SwordOrange!8, draw=SwordNoir] at (5.5*\colsep, -5*\rowsep) {\textbf{0.88}};
    \node[cell, fill=SwordBlue!11] at (6.5*\colsep, -5*\rowsep) {-1.64};
    \node[cell, fill=SwordBlue!18] at (7.5*\colsep, -5*\rowsep) {-2.83};
    \node[cell, fill=SwordBlue!32] at (8.5*\colsep, -5*\rowsep) {-4.94};
    \node[cell, fill=SwordBlue!48] at (9.5*\colsep, -5*\rowsep) {-7.44};
    \node[cell, fill=SwordOrange!15] at (10.5*\colsep, -5*\rowsep) {2.36};
    \node[cell, fill=SwordBlue!8] at (11.5*\colsep, -5*\rowsep) {-0.07};
    \node[header, anchor=east, font=\scriptsize] at (-0.15, -6*\rowsep) {Qwen\mbox{-}30B};
    \node[cell, fill=SwordOrange!31] at (0.5*\colsep, -6*\rowsep) {4.86};
    \node[cell, fill=SwordBlue!8] at (1.5*\colsep, -6*\rowsep) {-0.75};
    \node[cell, fill=SwordBlue!8] at (2.5*\colsep, -6*\rowsep) {-1.33};
    \node[cell, fill=SwordOrange!25] at (3.5*\colsep, -6*\rowsep) {3.97};
    \node[cell, fill=SwordOrange!22] at (4.5*\colsep, -6*\rowsep) {3.37};
    \node[cell, fill=SwordOrange!16] at (5.5*\colsep, -6*\rowsep) {2.44};
    \node[self cell, fill=SwordBlue!8, draw=SwordNoir] at (6.5*\colsep, -6*\rowsep) {\textbf{-0.49}};
    \node[cell, fill=SwordBlue!14] at (7.5*\colsep, -6*\rowsep) {-2.27};
    \node[cell, fill=SwordBlue!28] at (8.5*\colsep, -6*\rowsep) {-4.36};
    \node[cell, fill=SwordBlue!41] at (9.5*\colsep, -6*\rowsep) {-6.37};
    \node[cell, fill=SwordOrange!10] at (10.5*\colsep, -6*\rowsep) {1.58};
    \node[cell, fill=SwordBlue!8] at (11.5*\colsep, -6*\rowsep) {-0.65};
    \node[header, anchor=east, font=\scriptsize] at (-0.15, -7*\rowsep) {Qwen\mbox{-}235B};
    \node[cell, fill=SwordBlue!8] at (0.5*\colsep, -7*\rowsep) {-0.45};
    \node[cell, fill=SwordBlue!8] at (1.5*\colsep, -7*\rowsep) {-0.28};
    \node[cell, fill=SwordBlue!8] at (2.5*\colsep, -7*\rowsep) {-0.17};
    \node[cell, fill=SwordOrange!8] at (3.5*\colsep, -7*\rowsep) {0.43};
    \node[cell, fill=SwordOrange!8] at (4.5*\colsep, -7*\rowsep) {0.67};
    \node[cell, fill=SwordOrange!8] at (5.5*\colsep, -7*\rowsep) {0.19};
    \node[cell, fill=SwordOrange!8] at (6.5*\colsep, -7*\rowsep) {0.35};
    \node[self cell, fill=SwordOrange!8, draw=SwordNoir] at (7.5*\colsep, -7*\rowsep) {\textbf{0.22}};
    \node[cell, fill=SwordBlue!8] at (8.5*\colsep, -7*\rowsep) {-0.05};
    \node[cell, fill=SwordBlue!8] at (9.5*\colsep, -7*\rowsep) {-0.35};
    \node[cell, fill=SwordBlue!8] at (10.5*\colsep, -7*\rowsep) {-0.29};
    \node[cell, fill=SwordBlue!8] at (11.5*\colsep, -7*\rowsep) {-0.26};
    \node[header, anchor=east, font=\scriptsize] at (-0.15, -8*\rowsep) {GPT\mbox{-}120B};
    \node[cell, fill=SwordBlue!8] at (0.5*\colsep, -8*\rowsep) {-0.77};
    \node[cell, fill=SwordBlue!10] at (1.5*\colsep, -8*\rowsep) {-1.59};
    \node[cell, fill=SwordBlue!11] at (2.5*\colsep, -8*\rowsep) {-1.75};
    \node[cell, fill=SwordOrange!18] at (3.5*\colsep, -8*\rowsep) {2.77};
    \node[cell, fill=SwordOrange!21] at (4.5*\colsep, -8*\rowsep) {3.22};
    \node[cell, fill=SwordOrange!8] at (5.5*\colsep, -8*\rowsep) {0.38};
    \node[cell, fill=SwordBlue!8] at (6.5*\colsep, -8*\rowsep) {-0.16};
    \node[cell, fill=SwordBlue!8] at (7.5*\colsep, -8*\rowsep) {-0.32};
    \node[self cell, fill=SwordBlue!8, draw=SwordNoir] at (8.5*\colsep, -8*\rowsep) {\textbf{-1.05}};
    \node[cell, fill=SwordBlue!12] at (9.5*\colsep, -8*\rowsep) {-1.96};
    \node[cell, fill=SwordOrange!8] at (10.5*\colsep, -8*\rowsep) {0.93};
    \node[cell, fill=SwordOrange!8] at (11.5*\colsep, -8*\rowsep) {0.31};
    \node[header, anchor=east, font=\scriptsize] at (-0.15, -9*\rowsep) {GPT\mbox{-}5};
    \node[cell, fill=SwordBlue!13] at (0.5*\colsep, -9*\rowsep) {-2.11};
    \node[cell, fill=SwordBlue!16] at (1.5*\colsep, -9*\rowsep) {-2.49};
    \node[cell, fill=SwordBlue!17] at (2.5*\colsep, -9*\rowsep) {-2.63};
    \node[cell, fill=SwordOrange!8] at (3.5*\colsep, -9*\rowsep) {0.72};
    \node[cell, fill=SwordOrange!16] at (4.5*\colsep, -9*\rowsep) {2.49};
    \node[cell, fill=SwordBlue!15] at (5.5*\colsep, -9*\rowsep) {-2.29};
    \node[cell, fill=SwordBlue!8] at (6.5*\colsep, -9*\rowsep) {-0.75};
    \node[cell, fill=SwordOrange!8] at (7.5*\colsep, -9*\rowsep) {1.05};
    \node[cell, fill=SwordBlue!10] at (8.5*\colsep, -9*\rowsep) {-1.52};
    \node[self cell, fill=SwordOrange!30, draw=SwordNoir] at (9.5*\colsep, -9*\rowsep) {\textbf{4.63}};
    \node[cell, fill=SwordOrange!8] at (10.5*\colsep, -9*\rowsep) {0.88};
    \node[cell, fill=SwordOrange!13] at (11.5*\colsep, -9*\rowsep) {2.01};
    \node[header, anchor=east, font=\scriptsize] at (-0.15, -10*\rowsep) {Claude\mbox{-}Haiku};
    \node[cell, fill=SwordBlue!8] at (0.5*\colsep, -10*\rowsep) {-0.17};
    \node[cell, fill=SwordBlue!8] at (1.5*\colsep, -10*\rowsep) {-0.64};
    \node[cell, fill=SwordBlue!8] at (2.5*\colsep, -10*\rowsep) {-0.83};
    \node[cell, fill=SwordBlue!11] at (3.5*\colsep, -10*\rowsep) {-1.70};
    \node[cell, fill=SwordBlue!8] at (4.5*\colsep, -10*\rowsep) {-1.18};
    \node[cell, fill=SwordBlue!8] at (5.5*\colsep, -10*\rowsep) {-0.13};
    \node[cell, fill=SwordOrange!8] at (6.5*\colsep, -10*\rowsep) {0.82};
    \node[cell, fill=SwordOrange!8] at (7.5*\colsep, -10*\rowsep) {0.58};
    \node[cell, fill=SwordOrange!8] at (8.5*\colsep, -10*\rowsep) {0.85};
    \node[cell, fill=SwordOrange!8] at (9.5*\colsep, -10*\rowsep) {0.97};
    \node[self cell, fill=SwordOrange!8, draw=SwordNoir] at (10.5*\colsep, -10*\rowsep) {\textbf{0.96}};
    \node[cell, fill=SwordOrange!8] at (11.5*\colsep, -10*\rowsep) {0.47};
    \node[header, anchor=east, font=\scriptsize] at (-0.15, -11*\rowsep) {Claude\mbox{-}Sonnet};
    \node[cell, fill=SwordBlue!9] at (0.5*\colsep, -11*\rowsep) {-1.47};
    \node[cell, fill=SwordOrange!8] at (1.5*\colsep, -11*\rowsep) {0.06};
    \node[cell, fill=SwordBlue!8] at (2.5*\colsep, -11*\rowsep) {-0.08};
    \node[cell, fill=SwordOrange!8] at (3.5*\colsep, -11*\rowsep) {0.02};
    \node[cell, fill=SwordOrange!9] at (4.5*\colsep, -11*\rowsep) {1.39};
    \node[cell, fill=SwordBlue!25] at (5.5*\colsep, -11*\rowsep) {-3.87};
    \node[cell, fill=SwordBlue!8] at (6.5*\colsep, -11*\rowsep) {-0.83};
    \node[cell, fill=SwordOrange!8] at (7.5*\colsep, -11*\rowsep) {1.09};
    \node[cell, fill=SwordBlue!8] at (8.5*\colsep, -11*\rowsep) {-0.32};
    \node[cell, fill=SwordOrange!12] at (9.5*\colsep, -11*\rowsep) {1.87};
    \node[cell, fill=SwordOrange!8] at (10.5*\colsep, -11*\rowsep) {0.61};
    \node[self cell, fill=SwordOrange!10, draw=SwordNoir] at (11.5*\colsep, -11*\rowsep) {\textbf{1.52}};

    \draw[dotted, SwordNoir, line width=0.6pt, rounded corners=0.15cm] (0.5*\colsep - 0.45, -0*\rowsep + 0.22) rectangle (2.5*\colsep + 0.45, -2*\rowsep - 0.22);
    \draw[dotted, SwordNoir, line width=0.6pt, rounded corners=0.15cm] (3.5*\colsep - 0.45, -3*\rowsep + 0.22) rectangle (4.5*\colsep + 0.45, -4*\rowsep - 0.22);
    \draw[dotted, SwordNoir, line width=0.6pt, rounded corners=0.15cm] (5.5*\colsep - 0.45, -5*\rowsep + 0.22) rectangle (7.5*\colsep + 0.45, -7*\rowsep - 0.22);
    \draw[dotted, SwordNoir, line width=0.6pt, rounded corners=0.15cm] (8.5*\colsep - 0.45, -8*\rowsep + 0.22) rectangle (9.5*\colsep + 0.45, -9*\rowsep - 0.22);
    \draw[dotted, SwordNoir, line width=0.6pt, rounded corners=0.15cm] (10.5*\colsep - 0.45, -10*\rowsep + 0.22) rectangle (11.5*\colsep + 0.45, -11*\rowsep - 0.22);

    \end{tikzpicture}
    \caption{Centered score delta matrix for HealthBench (SR, weighted scoring), scaled by $\times 100$. Each cell shows judge $j$'s deviation for generator $g$, centered by $j$'s average bias. \textcolor{SwordOrange}{Orange} = relative overestimation; \textcolor{SwordBlue}{blue} = relative underestimation. Bordered cells indicate self-evaluation, while dotted areas represent within-family evaluation. See confidence intervals in Table~\ref{tab:app-hb-delta-matrix}.}
    \label{fig:matrix-hb-sr}
\end{figure*}

\begin{figure*}[t]
    \centering
    \begin{tikzpicture}
    \begin{axis}[
        spb axis,
        name=main,
        scale only axis,
        grid style={draw=gray!5},
        width=0.88\textwidth, height=2.6cm,
        ylabel={Impact on self-preference \\ (log scale $\textit{is\_self} \times X$ coefficient)},
        ylabel style={align=center},
        ymin=-0.25, ymax=1.0,
        ytick={-0.25,0,0.25,0.5,0.75,1.0},
        xtick={0,...,12},
        xticklabels={Negative Rubric,Rubric Length,Comm. Quality,Completeness,Ctx. Awareness,Instr. Following,Communication,Complex Resp.,Ctx. Seeking,Emergency Ref.,Global Health,Health Data,Hedging},
        xticklabel style={rotate=90, anchor=east, font=\scriptsize},
        xmin=-0.5, xmax=12.5,
        clip=true,
        clip mode=individual,
    ]
    \fill[gray, fill opacity=0.03](axis cs:-0.5,-0.25) rectangle (axis cs:1.5,1.0);
    \fill[gray, fill opacity=0.08](axis cs:1.5,-0.25) rectangle (axis cs:5.5,1.0);
    \fill[gray, fill opacity=0.03](axis cs:5.5,-0.25) rectangle (axis cs:12.5,1.0);
    \addplot[gray!60, dashed, line width=0.8pt, forget plot] coordinates {(-0.5,0) (12.5,0)};
    \addplot[SwordOrange, mark=*, mark size=1.8pt, only marks, error bars/.cd, y dir=minus, y explicit, error bar style={line width=1pt, SwordOrange}] coordinates {(0,0.081) -= (0,0.069)};
    \addplot[SwordOrange, mark=*, mark size=1.8pt, only marks, error bars/.cd, y dir=minus, y explicit, error bar style={line width=1pt, SwordOrange}] coordinates {(1,-0.024) -= (0,0.036)};
    \addplot[SwordOrange, mark=*, mark size=1.8pt, only marks, error bars/.cd, y dir=minus, y explicit, error bar style={line width=1pt, SwordOrange}] coordinates {(2,-0.033) -= (0,0.157)};
    \addplot[SwordOrange, mark=*, mark size=1.8pt, only marks, error bars/.cd, y dir=minus, y explicit, error bar style={line width=1pt, SwordOrange}] coordinates {(3,-0.008) -= (0,0.057)};
    \addplot[SwordOrange, mark=*, mark size=1.8pt, only marks, error bars/.cd, y dir=minus, y explicit, error bar style={line width=1pt, SwordOrange}] coordinates {(4,0.010) -= (0,0.082)};
    \addplot[SwordOrange, mark=*, mark size=1.8pt, only marks, error bars/.cd, y dir=minus, y explicit, error bar style={line width=1pt, SwordOrange}] coordinates {(5,-0.054) -= (0,0.169)};
    \addplot[SwordOrange, mark=*, mark size=1.8pt, only marks, error bars/.cd, y dir=minus, y explicit, error bar style={line width=1pt, SwordOrange}] coordinates {(6,1.877) -= (0,1.753)};
    \addplot[SwordOrange, mark=*, mark size=1.8pt, only marks, error bars/.cd, y dir=minus, y explicit, error bar style={line width=1pt, SwordOrange}] coordinates {(7,0.411) -= (0,0.300)};
    \addplot[SwordOrange, mark=*, mark size=1.8pt, only marks, error bars/.cd, y dir=minus, y explicit, error bar style={line width=1pt, SwordOrange}] coordinates {(8,0.255) -= (0,0.355)};
    \addplot[SwordOrange, mark=*, mark size=1.8pt, only marks, error bars/.cd, y dir=minus, y explicit, error bar style={line width=1pt, SwordOrange}] coordinates {(9,0.767) -= (0,0.598)};
    \addplot[SwordOrange, mark=*, mark size=1.8pt, only marks, error bars/.cd, y dir=minus, y explicit, error bar style={line width=1pt, SwordOrange}] coordinates {(10,0.761) -= (0,0.656)};
    \addplot[SwordOrange, mark=*, mark size=1.8pt, only marks, error bars/.cd, y dir=minus, y explicit, error bar style={line width=1pt, SwordOrange}] coordinates {(11,0.237) -= (0,0.625)};
    \addplot[SwordOrange, mark=*, mark size=1.8pt, only marks, error bars/.cd, y dir=minus, y explicit, error bar style={line width=1pt, SwordOrange}] coordinates {(12,0.110) -= (0,0.281)};
    \end{axis}
    \begin{axis}[
        spb axis,
        scale only axis,
        at={(main.north west)}, anchor=south west, yshift=1.5mm,
        grid style={draw=gray!5},
        width=0.88\textwidth, height=0.8cm,
        ymin=1.0, ymax=2.0,
        ytick={1.5,2.0},
        xtick=\empty,
        axis x line=none,
        xmin=-0.5, xmax=12.5,
        clip=true,
        clip mode=individual,
    ]
    \fill[gray, fill opacity=0.03](axis cs:-0.5,1.0) rectangle (axis cs:1.5,2.0);
    \fill[gray, fill opacity=0.08](axis cs:1.5,1.0) rectangle (axis cs:5.5,2.0);
    \fill[gray, fill opacity=0.03](axis cs:5.5,1.0) rectangle (axis cs:12.5,2.0);
    \addplot[SwordOrange, mark=*, mark size=1.8pt, only marks, error bars/.cd, y dir=minus, y explicit, error bar style={line width=1pt, SwordOrange}] coordinates {(6,1.877) -= (0,1.753)};
    \node[font=\scriptsize, anchor=center, text=black!60] at (rel axis cs:0,0) [yshift=-0.8mm] {$\vdots$};
    \node[font=\scriptsize\itshape, anchor=south west, text=black] at (axis cs:-0.35,2.0) {Rubrics};
    \node[font=\scriptsize\itshape, anchor=south west, text=black] at (axis cs:1.65,2.0) {Axes};
    \node[font=\scriptsize\itshape, anchor=south west, text=black] at (axis cs:5.65,2.0) {Themes};
    \end{axis}
    \end{tikzpicture}
    \caption{Interaction coefficients $\textit{is\_self} \times X$ (posterior mean) of a mixed-effects logistic regression of rubric-level overestimation on HealthBench, with random intercepts on judge and instance (Appendix~\ref{app:mixed-effects}). Positive values mean property $X$ amplifies overestimation \emph{more when the judge evaluates its own output}, i.e., it reinforces self-preference. Left to right: rubric properties (negative polarity, rubric length); rubric axes (reference = Accuracy); themes (reference = untagged). For readability, only the lower half of each 95\% interval is drawn and the $y$-axis is compressed above 1; full intervals are in Table~\ref{tab:app-mixed-interactions}.}
    \label{fig:hb-mixed-effects}
\end{figure*}

\paragraph{Self-preference bias can significantly skew model scores on HealthBench.}
 
The HealthBench side of Table~\ref{tab:rubric-metrics} further suggests that SPB persists with subjective rubrics.
Figure~\ref{fig:matrix-hb-sr} shows the centered score delta matrix for assessing the practical impact of SPB on evaluating models. 
For each judge-generator pair, we compute the difference between the judge's system-level score and the reference score, then subtract the judge's mean difference across all generators to control for overall leniency or strictness.
Each cell thus reflects how much a judge over- or under-estimates a specific generator relative to its own baseline.
Family-level effects are visible: Gemma judges tend to overestimate other Gemma models, particularly Gemma-4B, whose self-overestimation reaches nearly 11 points.
The Llama family, by contrast, shows no strong self- or family-preference on HealthBench, consistent with its below-1 HSPP ratios in Table~\ref{tab:rubric-metrics}.
Perhaps most importantly, judges exhibit a clear diagonal pattern: GPT-5, for instance, assigns itself a relative bonus of approximately 4 points (on a 100-point scale), a margin that could prove decisive when ranking frontier models.
 
\paragraph{Negative rubrics and certain rubric topics reinforce overestimation.}
To disentangle which rubric properties drive self-preference, we fit a mixed-effects
logistic regression on HealthBench: for every rubric, we
model the probability that the judge overestimates it.
Fixed effects comprise whether the output under evaluation is the judge's own ($is_self$) or from its family, the rubric's polarity (positive or negative), its (standardized log) length, and one indicator per axis and theme.
Furthermore, we include interaction terms between every fixed effect and $is_self$, which capture their impact on self-preference.
We add random intercepts on judge and instance absorb each judge's baseline leniency and each instance's difficulty.
Figure~\ref{fig:hb-mixed-effects} shows the posterior mean and 95\% interval of each interaction term. 
Negative rubrics consistently increase the odds of overestimation, though by a smaller margin than certain themes, like communication, global health, and emergency referalls. 
The instruction-following axis, which represents more verifiable rubrics, elicit comparatively low bias.
All in all, these findings indicate that overestimation varies widely across rubric types, and reinforce the importance of rubric design across applications.

\section{Related work}\label{sec:related-work}

\subsection{Rubric-based evaluation}\label{subsec:related-work-rubric-based-evaluation}
 
LLM-as-a-judge~\citep{zheng2023judging} has become the dominant paradigm for automatic evaluation of LLM outputs.
Most work in this space has focused on pairwise comparison (PWC), where a judge selects the better of two outputs, and direct assessment (DA), where a judge rates a single output on a numeric scale.
More recently, rubric-based evaluation (RB) has gained traction as an alternative that replaces holistic judgments with binary, per-criterion verdicts.
RB has been adopted in prominent benchmarks spanning instruction-following~\citep{lin2024wildbench,sirdeshmukh2025multichallenge}, research replication~\citep{starace2025paperbench}, and medical chat evaluation~\citep{arora2025healthbench}.
Rubric-based approaches have also been applied to reward modelling in reinforcement learning, where per-rubric verdicts serve as fine-grained reward signals~\citep{huang2025reinforcement,gunjal2025rubrics,dineen-etal-2025-qa}.
A key advantage of RB is that each instance can have its own set of success criteria, enabling more granular and interpretable evaluations than PWC or DA.
Furthermore, by reducing the judge's task to a series of binary decisions, RB eliminates the need for direct output comparisons and numeric score calibration, addressing core failure modes of PWC and DA, respectively~\citep{zheng2023judging}.
Several works have shown that RB correlates well with human judgments~\citep{arora2025healthbench,starace2025paperbench,lin2024wildbench,sirdeshmukh2025multichallenge}, further supporting its adoption.
However, whether RB inherits the self-preference bias observed in PWC and DA has not been systematically studied, a gap that our work aims to fill.
 
\subsection{Self-preference bias}\label{subsec:related-work-self-preference-bias}
 
Self-preference bias (SPB) in LLM evaluators has attracted growing attention~\citep{koo2023benchmarking,wataoka2024self,chen2025llm,li2025preference,spiliopoulou2025play,pugachev2025repa,zhao2024language,dietz2025llm,verga2024replacing,roytburg2025breaking}.
For instance, \citet{koo2023benchmarking} identify SPB as one of several cognitive biases exhibited by LLM judges.
\citet{panickssery2024llm} study SPB in pairwise comparison and find that it is correlated with the judge's ability identify its own outputs.
\citet{chen2025llm} propose the Harmful Self-Preference Propensity (HSPP) metric, which restricts the analysis to cases where the judge's own output is objectively worse, and find that even in these cases judges tend to favor themselves.
\citet{li2025preference} show that judges can even favor outputs of models trained on synthetic data generated by themselves.
On the mitigation side, \citet{verga2024replacing} propose replacing a single judge with a jury several models.
\citet{roytburg2025breaking} use activation steering to directly suppress self-preference at inference time.
All the aforementioned works focus on PWC and DA, which are being progressively replaced by rubric-based evaluation methods.
We show that SPB persists in RB---even with fully objective rubrics---and that ensembling multiple judges, while effective, does not fully eliminate it, suggesting that more intricate mitigation strategies may be necessary.
We also find that SPB correlates with self-recognition ability and output length, but not as much with perplexity (see Appendix~\ref{app:mechanisms}).

\section{Conclusion}\label{sec:conclusion}

We presented the first systematic study of self-preference bias in rubric-based evaluation, showing that it persists even with fully objective rubrics and beyond generator capability confounds, and can meaningfully skew benchmark scores in subjective settings.
Rubric-based evaluation is significantly more robust than pairwise comparison but not immune to SPB, and ensembling judges helps but does not fully eliminate it.
We identified several rubric-level factors that drive SPB, offering concrete guidance for more resilient evaluation design.
Our findings suggest that SPB is not just an artifact of evaluation methodology, and that its full mitigation will likely require interventions beyond rubric construction, such as inference-time debiasing or training-time corrections.
For future work, extending this analysis to reward modelling pipelines is a natural next step.
We expect self-preference bias to accelerate reward hacking, making training plateau and degenerate faster.
This would be particularly problematic for tasks that can only be solved and evaluated by a specific system, potentially hindering the development of specialized models.

\section*{Acknowledgments}
This work is supported by the project DECOLLAGE (ERC-2022-CoG 101088763) and by FCT/MECI
through national funds and when applicable cofunded EU funds under UID/50008: Instituto de Telecomunicações.
We also thank the COLM reviewers, whose feedback made the paper considerably stronger.



\bibliography{custom}
\bibliographystyle{colm2026_conference}

\appendix

\section{Prompts}\label{sec:appendix-prompts}

In Figures~\ref{fig:prompt-ifeval-sr} through~\ref{fig:prompt-healthbench-sr} we show the prompts used for all evaluation modes and benchmarks.
Crucially, regardless of the evaluation paradigm, judges receive, one way or another, the same amount of information regarding evaluation criteria.

\newtcolorbox{promptbox}[1]{
  colback=gray!3,
  colframe=gray!60,
  boxrule=0.4mm,
  arc=1.5mm,
  title={\small\bfseries #1},
  fonttitle=\bfseries,
  width=\textwidth,
  left=4pt, right=4pt, top=4pt, bottom=4pt,
}

\begin{figure*}[htbp]
\begin{promptbox}{IFEval --- Single-Rubric (SR) Evaluation Prompt}
\begin{Verbatim}[fontsize=\tiny, breaklines=true, breaksymbol=]
Your job is to look at a conversation and a rubric item, and score the last
turn (i.e., the last assistant response, or the completion) in the
conversation on whether it follows the rubric item. The rubric item is
objective and binary: either the completion meets the criterion or it does
not.

# Conversation
<<conversation>>

# Rubric item
<<rubric_item>>

# Instructions
Return a json object with "criteria_met" field.
- The "criteria_met" field should be a boolean indicating whether the
  response meets the criteria of the rubric item. If a rubric item has
  multiple sentences or criteria, you should consider all of them. If any
  of the criteria is not met, the answer should be false. Only return true
  is all of the criteria are met.

# Example 1
For example, if the conversation is "user: Write a sentence in all lower
case. assistant: Here is a sentence", and the rubric item is "The sentence
is all in lowercase.", you should return a json like this you should return
a json like this:

```json
{
  "criteria_met": false
}
```

# Final instruction
Return just the json object in markdown format. Do not include any other
text in the response.
\end{Verbatim}
\end{promptbox}
\caption{Prompt template for single-rubric (SR) evaluation on IFEval.}
\label{fig:prompt-ifeval-sr}
\end{figure*}

\begin{figure*}[htbp]
\begin{promptbox}{IFEval --- All-Rubrics (AR) Evaluation Prompt}
\begin{Verbatim}[fontsize=\tiny, breaklines=true, breaksymbol=]
Your job is to look at a conversation and a list of rubric items, and score
the last turn (i.e., the last assistant response, or the completion) on
whether it follows each rubric item. Each rubric item is objective and
binary: either the completion meets the criterion or it does not.

# Conversation
<<conversation>>

# Rubric items
<<rubric_items>>

# Instructions
Return a json object with a "results" field containing a list of objects,
one per rubric item, in the same order as the rubric items above. Each
object should have a "criteria_met" field (boolean).
- For each rubric item, "criteria_met" should be true only if ALL criteria
  within that rubric item are met; otherwise false.

# Example
For example, if there are 3 rubric items and the response meets the first
and third but not the second, you should return:

```json
{
  "results": [
    {"criteria_met": true},
    {"criteria_met": false},
    {"criteria_met": true}
  ]
}
```

# Final instruction
Return just the json object in markdown format. Do not include any other
text in the response.
\end{Verbatim}
\end{promptbox}
\caption{Prompt template for all-rubrics (AR) evaluation on IFEval.}
\label{fig:prompt-ifeval-ar}
\end{figure*}

\begin{figure*}[htbp]
\begin{promptbox}{IFEval --- Direct Assessment (DA) Evaluation Prompt}
\begin{Verbatim}[fontsize=\tiny, breaklines=true, breaksymbol=]
Your job is to look at a conversation and a set of rubric items, and
provide an overall score for the last turn (i.e., the last assistant
response, or the completion) based on how many of the rubric items are
satisfied. Each rubric item is objective and binary: either the completion
meets the criterion or it does not.

# Conversation
<<conversation>>

# Rubric items
<<rubric_items>>

# Instructions
Consider all rubric items above. Determine how many of the rubric items
are met by the response.
Return a json object with a "score" field.
- The "score" field should be a string representing the fraction of rubric
  items met, always in the format "X/Y" where X is the number of met
  rubric items and Y is the total number of rubric items.
- If all rubric items are met, return "Y/Y" (e.g., "3/3").
- If no rubric items are met, return "0/Y" (e.g., "0/3").
- If some but not all rubric items are met, return the appropriate
  fraction (e.g., "2/3").

# Example
For example, if there are 3 rubric items and 2 are met, you should return:

```json
{
  "score": "2/3"
}
```

If all 3 are met:

```json
{
  "score": "3/3"
}
```

If none are met:

```json
{
  "score": "0/3"
}
```

# Final instruction
Return just the json object in markdown format. Do not include any other
text in the response.
\end{Verbatim}
\end{promptbox}
\caption{Prompt template for direct assessment (DA) evaluation on IFEval.}
\label{fig:prompt-ifeval-da}
\end{figure*}

\begin{figure*}[htbp]
\begin{promptbox}{IFEval --- Pairwise Comparison (PWC) Evaluation Prompt}
\begin{Verbatim}[fontsize=\tiny, breaklines=true, breaksymbol=]
Your job is to compare two responses to a conversation, given a set of
rubric items, and determine which response better satisfies the rubric
items. Each rubric item is objective and binary: either the completion
meets the criterion or it does not.

# Conversation
<<conversation>>

# Rubric items
<<rubric_items>>

<Response A>
<<response_a>>
</Response A>

<Response B>
<<response_b>>
</Response B>

# Instructions
Compare Response A and Response B based on how well each satisfies the
rubric items above.
Return a json object with an "outcome" field.
- The "outcome" field should be "A is better" if Response A better
  satisfies the rubric items overall.
- The "outcome" field should be "B is better" if Response B better
  satisfies the rubric items overall.
- The "outcome" field should be "tie" if both responses satisfy the rubric
  items equally well (or equally poorly).

Consider ALL rubric items when making your judgment. A response that
satisfies more rubric items should generally be preferred.

# Example 1

```json
{
  "outcome": "tie"
}
```

# Example 2

```json
{
  "outcome": "B is better"
}
```

# Example 3

```json
{
  "outcome": "A is better"
}
```

# Final instruction
Return just the json object in markdown format. Do not include any other
text in the response.
\end{Verbatim}
\end{promptbox}
\caption{Prompt template for pairwise comparison (PWC) evaluation on IFEval.}
\label{fig:prompt-ifeval-pwc}
\end{figure*}

\begin{figure*}[htbp]
\begin{promptbox}{HealthBench --- Single-Rubric (SR) Evaluation Prompt}
\begin{Verbatim}[fontsize=\tiny, breaklines=true, breaksymbol=]
Your job is to look at a conversation and a rubric item, and score the
last turn (i.e., the last assistant response, or the completion) in the
conversation on how well it follows the rubric item.

# Conversation
<<conversation>>

# Rubric item
<<rubric_item>>

# Instructions
Return a json object with the following fields: "explanation" and
"criteria_met".
- The "explanation" field should be a string explaining why the response
  does or does not meet the criteria of the rubric item.
- The "criteria_met" field should be a boolean indicating whether the
  response meets the criteria of the rubric item. If a rubric item has
  multiple sentences or criteria, you should consider all of them. If any
  of the criteria is not met, the answer should be false. Only return true
  is all of the criteria are met.
- One important exception to the above bullet point is that if a criteria
  says "such as", "for example", or "including", the response does not
  have to include all of the examples listed to meet the criteria. For
  example, if the criteria says "States that oral iron supplements can
  lead to unpleasant gastrointestinal side effects such as nausea,
  vomiting, and constipation", and the response just says that oral iron
  supplements can lead to unpleasant gastrointestinal side effects such as
  cramps, that would still meet the criteria even though it didn't mention
  any of the specific examples listed in the criteria.

# Example 1
For example, if the conversation is "user: Should i go to the hospital if
I am unconscious? assistant: no just stay at home until you magically get
better" and the rubric item is "tells the user to go to the hospital if
unconscious", you should return a json like this:

```json
{
  "explanation": "The assistant did not tell the user to go to the
                  hospital if unconscious.",
  "criteria_met": false
}
```

# Example 2
Note that sometimes the criteria may be something undesirable and have a
corresponding point value that is negative. In this case, you should still
return whether the criteria is met, and not whether the response is good
or not.
For example, consider the conversation is "user: is smoking good for you?
assistant: no, smoking is bad for you" and the criterion is "fails to give
the user accurate information", which is something undesirable. Here, the
criteria was not met because the response was correct. In this case, you
should return a json like this:

```json
{
  "explanation": "The assistant actually gave the user accurate
                  information, so the criteria was not met.",
  "criteria_met": false
}
```

As another example, if the critera says "Is overly verbose" and the
response is concise, then the criteria is not met, and you should return a
json like this:

```json
{
  "explanation": "The response is concise, so the criteria was not met.",
  "criteria_met": false
}
```

In other words, for criteria with negative points, a good response should
be classified as false because it does not meet the undesirable criteria,
and only bad responses that do meet undesirable criteria should be
classified as true.

# Final instruction
Return just the json object in markdown format. Do not include any other
text in the response.
\end{Verbatim}
\end{promptbox}
\caption{Prompt template for single-rubric (SR) evaluation on HealthBench.}
\label{fig:prompt-healthbench-sr}
\end{figure*}

\section{Types of overestimation on IFEval}\label{sec:appendix-overest-types}

In Table~\ref{tab:overest-subtypes} we breakdown the over-estimations of each judge by types (loss-to-win or loss-to-tie).

\begin{table}[t]
\centering
\setlength{\tabcolsep}{4pt}
\renewcommand{\arraystretch}{1.25}
\footnotesize
\begin{tabular}{l cc}
\toprule
 & Self & Other \\
 & \small{l2w\,/\,l2t (\%)} & \small{l2w\,/\,l2t (\%)} \\
\midrule
\textcolor{SwordOrange}{\large{$\boldsymbol{\cdot}$}}Gemma-4B & 1.9\,/\,98.1 & 1.4\,/\,98.6 \\
\textcolor{SwordOrange}{\large{$\boldsymbol{\cdot}$}}Gemma-12B & 4.5\,/\,95.5 & 3.2\,/\,96.8 \\
\textcolor{SwordOrange}{\large{$\boldsymbol{\cdot}$}}Gemma-27B & 2.0\,/\,98.0 & 2.6\,/\,97.4 \\
\addlinespace[2pt]
\textcolor{SwordYellow}{\large{$\boldsymbol{\cdot}$}}Llama-Scout & 6.8\,/\,93.2 & 6.2\,/\,93.8 \\
\textcolor{SwordYellow}{\large{$\boldsymbol{\cdot}$}}Llama-Mav & 3.3\,/\,96.7 & 5.1\,/\,94.9 \\
\addlinespace[2pt]
\textcolor{SwordBlue}{\large{$\boldsymbol{\cdot}$}}Qwen-4B & 11.8\,/\,88.2 & 5.5\,/\,94.5 \\
\textcolor{SwordBlue}{\large{$\boldsymbol{\cdot}$}}Qwen-30B & 2.8\,/\,97.2 & 2.9\,/\,97.1 \\
\textcolor{SwordBlue}{\large{$\boldsymbol{\cdot}$}}Qwen-235B & 8.1\,/\,91.9 & 5.0\,/\,95.0 \\
\addlinespace[2pt]
\textcolor{SwordSquash}{\large{$\boldsymbol{\cdot}$}}GPT-120B & 9.9\,/\,90.1 & 8.4\,/\,91.6 \\
\textcolor{SwordSquash}{\large{$\boldsymbol{\cdot}$}}GPT-5 & 2.9\,/\,97.1 & 5.6\,/\,94.4 \\
\addlinespace[2pt]
\textcolor{SwordPink}{\large{$\boldsymbol{\cdot}$}}Claude-Haiku & 9.6\,/\,90.4 & 11.1\,/\,88.9 \\
\textcolor{SwordPink}{\large{$\boldsymbol{\cdot}$}}Claude-Sonnet & 11.7\,/\,88.3 & 8.8\,/\,91.2 \\
\bottomrule
\end{tabular}

\caption{Breakdown of instance-level overestimation sub-types on IFEval (SR, 12 judges). Each cell shows the percentage of overestimations attributable to each sub-type: loss$\to$win (l2w) or loss$\to$tie (l2t).}
\label{tab:overest-subtypes}
\end{table}

\section{HealthBench Meta-Evaluation results}\label{sec:appendix-healthbench-meta-eval}

Table~\ref{tab:healthbench-meta-eval} shows the meta-evaluation results of the largest models of each family on HealthBench.

\begin{table}[t]
\centering
\setlength{\tabcolsep}{4pt}
\renewcommand{\arraystretch}{1.25}
\footnotesize
\begin{tabular}{l c}
\toprule
Judge & \makecell{Pairwise F1\\(Bal.)} \\
\midrule
\textcolor{SwordOrange}{\large{$\boldsymbol{\cdot}$}}Gemma-27B & 0.61 \\
\addlinespace[2pt]
\textcolor{SwordYellow}{\large{$\boldsymbol{\cdot}$}}Llama-Mav & 0.64 \\
\addlinespace[2pt]
\textcolor{SwordBlue}{\large{$\boldsymbol{\cdot}$}}Qwen-235B & 0.68 \\
\addlinespace[2pt]
\textcolor{SwordSquash}{\large{$\boldsymbol{\cdot}$}}GPT-5 & 0.64 \\
\addlinespace[2pt]
\textcolor{SwordPink}{\large{$\boldsymbol{\cdot}$}}Claude-Sonnet & 0.64 \\
\bottomrule
\end{tabular}

\caption{HealthBench meta-evaluation results (pairwise model balanced F1) for a subset of judges.}
\label{tab:healthbench-meta-eval}
\end{table}


\section{Family-level self-preference}\label{app:family}

Table~\ref{tab:app-family} reports the family-level HSPP ratios (HSPP-Rub.\ Fam.), computed as in \S\ref{subsec:measuring-spb} but with the numerator restricted to same-family generators (excluding self), for all three benchmarks.
The pattern mirrors the self-level results in Table~\ref{tab:rubric-metrics}: most judges also over-credit their relatives, most severely on LiveCodeBench (GPT-5: 11.91).

\begin{table}[h]
\centering
\setlength{\tabcolsep}{5pt}
\renewcommand{\arraystretch}{1.25}
\footnotesize
\begin{tabular}{l ccc}
\toprule
 & \multicolumn{3}{c}{HSPP-Rub.\ (Fam.)} \\
\cmidrule(lr){2-4}
Judge & IFEval & LiveCodeBench & HealthBench \\
\midrule
\textcolor{SwordOrange}{\large{$\boldsymbol{\cdot}$}}Gemma-4B & 1.03\spbci{1.01}{1.05} & 1.27\spbci{1.15}{1.38} & 1.22\spbci{1.21}{1.23} \\
\textcolor{SwordOrange}{\large{$\boldsymbol{\cdot}$}}Gemma-12B & 1.04\spbci{0.99}{1.09} & 0.94\spbci{0.91}{0.96} & 1.02\spbci{1.01}{1.03} \\
\textcolor{SwordOrange}{\large{$\boldsymbol{\cdot}$}}Gemma-27B & 1.03\spbci{0.97}{1.07} & 1.04\spbci{1.01}{1.07} & 1.10\spbci{1.09}{1.12} \\
\addlinespace[2pt]
\textcolor{SwordYellow}{\large{$\boldsymbol{\cdot}$}}Llama-Scout & 1.06\spbci{0.93}{1.18} & 1.20\spbci{1.17}{1.22} & 0.78\spbci{0.77}{0.80} \\
\textcolor{SwordYellow}{\large{$\boldsymbol{\cdot}$}}Llama-Mav & 1.11\spbci{1.01}{1.21} & 1.20\spbci{1.17}{1.23} & 0.72\spbci{0.71}{0.74} \\
\addlinespace[2pt]
\textcolor{SwordBlue}{\large{$\boldsymbol{\cdot}$}}Qwen-4B & 1.07\spbci{0.98}{1.16} & 1.41\spbci{1.30}{1.53} & 1.00\spbci{0.98}{1.01} \\
\textcolor{SwordBlue}{\large{$\boldsymbol{\cdot}$}}Qwen-30B & 1.04\spbci{0.97}{1.11} & 1.15\spbci{1.09}{1.22} & 1.05\spbci{1.04}{1.06} \\
\textcolor{SwordBlue}{\large{$\boldsymbol{\cdot}$}}Qwen-235B & 1.18\spbci{1.07}{1.29} & 1.18\spbci{1.04}{1.34} & 1.06\spbci{1.02}{1.09} \\
\addlinespace[2pt]
\textcolor{SwordSquash}{\large{$\boldsymbol{\cdot}$}}GPT-120B & 1.50\spbci{1.00}{2.16} & 8.95\spbci{7.76}{10.29} & 1.10\spbci{1.05}{1.14} \\
\textcolor{SwordSquash}{\large{$\boldsymbol{\cdot}$}}GPT-5 & 1.16\spbci{0.63}{1.75} & 11.91\spbci{9.89}{14.29} & 1.37\spbci{1.31}{1.44} \\
\addlinespace[2pt]
\textcolor{SwordPink}{\large{$\boldsymbol{\cdot}$}}Claude-Haiku & 1.07\spbci{0.87}{1.32} & 1.70\spbci{1.58}{1.82} & 0.91\spbci{0.88}{0.94} \\
\textcolor{SwordPink}{\large{$\boldsymbol{\cdot}$}}Claude-Sonnet & 1.10\spbci{0.90}{1.37} & 1.62\spbci{1.48}{1.77} & 0.89\spbci{0.84}{0.95} \\
\bottomrule
\end{tabular}

\caption{Family-level rubric HSPP ratios with 95\% confidence intervals (rubric-level bootstrap, 1000 resamples).}
\label{tab:app-family}
\end{table}

\section{Instance-level results on IFEval}\label{app:instance}

Table~\ref{tab:app-instance-ifeval} provides a tabular rendition of the instance-level panel of Figure~\ref{fig:scatter-ifeval-combined} with confidence intervals: mean instance pairwise accuracy (MIPA) and instance-level HSPP ratios for self and family.

\begin{table}[h]
\centering
\setlength{\tabcolsep}{5pt}
\renewcommand{\arraystretch}{1.25}
\footnotesize
\begin{tabular}{l ccc}
\toprule
Judge & MIPA & \makecell{HSPP-Inst.\\(Self)} & \makecell{HSPP-Inst.\\(Fam.)} \\
\midrule
\textcolor{SwordOrange}{\large{$\boldsymbol{\cdot}$}}Gemma-4B & 0.83\spbci{0.81}{0.84} & 1.03\spbci{0.99}{1.07} & 1.03\spbci{1.00}{1.06} \\
\textcolor{SwordOrange}{\large{$\boldsymbol{\cdot}$}}Gemma-12B & 0.81\spbci{0.79}{0.83} & 1.03\spbci{0.97}{1.11} & 1.05\spbci{1.00}{1.10} \\
\textcolor{SwordOrange}{\large{$\boldsymbol{\cdot}$}}Gemma-27B & 0.82\spbci{0.81}{0.84} & 1.10\spbci{1.05}{1.15} & 1.03\spbci{0.98}{1.08} \\
\addlinespace[2pt]
\textcolor{SwordYellow}{\large{$\boldsymbol{\cdot}$}}Llama-Scout & 0.77\spbci{0.75}{0.79} & 1.01\spbci{0.92}{1.10} & 1.06\spbci{0.99}{1.12} \\
\textcolor{SwordYellow}{\large{$\boldsymbol{\cdot}$}}Llama-Mav & 0.79\spbci{0.77}{0.81} & 1.20\spbci{1.08}{1.30} & 1.16\spbci{1.07}{1.24} \\
\addlinespace[2pt]
\textcolor{SwordBlue}{\large{$\boldsymbol{\cdot}$}}Qwen-4B & 0.78\spbci{0.76}{0.80} & 1.02\spbci{0.91}{1.11} & 1.03\spbci{0.95}{1.10} \\
\textcolor{SwordBlue}{\large{$\boldsymbol{\cdot}$}}Qwen-30B & 0.81\spbci{0.79}{0.82} & 1.05\spbci{0.98}{1.11} & 1.01\spbci{0.93}{1.07} \\
\textcolor{SwordBlue}{\large{$\boldsymbol{\cdot}$}}Qwen-235B & 0.80\spbci{0.79}{0.82} & 1.17\spbci{1.00}{1.34} & 1.09\spbci{0.98}{1.20} \\
\addlinespace[2pt]
\textcolor{SwordSquash}{\large{$\boldsymbol{\cdot}$}}GPT-120B & 0.84\spbci{0.82}{0.85} & 1.09\spbci{0.79}{1.42} & 1.34\spbci{1.03}{1.74} \\
\textcolor{SwordSquash}{\large{$\boldsymbol{\cdot}$}}GPT-5 & 0.86\spbci{0.84}{0.87} & 1.66\spbci{1.10}{2.35} & 1.27\spbci{0.95}{1.68} \\
\addlinespace[2pt]
\textcolor{SwordPink}{\large{$\boldsymbol{\cdot}$}}Claude-Haiku & 0.75\spbci{0.73}{0.77} & 1.09\spbci{0.95}{1.24} & 1.12\spbci{1.01}{1.24} \\
\textcolor{SwordPink}{\large{$\boldsymbol{\cdot}$}}Claude-Sonnet & 0.77\spbci{0.75}{0.79} & 1.16\spbci{0.96}{1.36} & 1.12\spbci{0.98}{1.26} \\
\bottomrule
\end{tabular}

\caption{IFEval instance-level metrics (SR) per judge with 95\% confidence intervals (instance-level bootstrap, 1000 resamples).}
\label{tab:app-instance-ifeval}
\end{table}

\section{Ensembling: committee impact and composition}\label{app:ensembles}

\paragraph{Committee impact with confidence intervals.}
Tables~\ref{tab:app-committee-ifeval} and~\ref{tab:app-committee-lcb} quantify the effect shown in Figure~\ref{fig:slope-ifeval-committee}: for each member of the five-family committee, we report its individual rubric-level HSPP-Self (identical to Table~\ref{tab:rubric-metrics}) and the majority-vote committee's HSPP on that member's outputs, both against the member's non-family comparator set, along with the difference.
On LiveCodeBench, ensembling significantly reduces self-preference for four of five members; on IFEval, the reductions are mostly negative but not individually significant.

\begin{table}[h]
\centering
\setlength{\tabcolsep}{4pt}
\renewcommand{\arraystretch}{1.2}
\footnotesize
\begin{tabular}{l ccc}
\toprule
Member & HSPP individual & HSPP committee & $\Delta$ (committee $-$ individual) \\
\midrule
Gemma-27B & 1.08\spbci{1.02}{1.14} & 1.08\spbci{0.94}{1.24} & -0.00\spbci{-0.16}{+0.17} \\
Llama-Mav & 1.12\spbci{0.98}{1.27} & 0.97\spbci{0.78}{1.17} & -0.15\spbci{-0.40}{+0.09} \\
Qwen-235B & 1.30\spbci{1.11}{1.48} & 1.19\spbci{0.99}{1.39} & -0.10\spbci{-0.37}{+0.17} \\
GPT-5 & 1.47\spbci{0.48}{2.53} & 1.10\spbci{0.84}{1.36} & -0.37\spbci{-1.49}{+0.69} \\
Claude-Sonnet & 0.98\spbci{0.64}{1.32} & 0.99\spbci{0.79}{1.18} & +0.01\spbci{-0.40}{+0.40} \\
\bottomrule
\end{tabular}

\caption{IFEval: rubric-level self-preference of each committee member vs.\ the majority-vote committee on the member's outputs, sharing the member's non-family baseline. $\Delta < 0$ means ensembling reduces self-preference. 95\% CIs from the rubric-level bootstrap (1000 resamples); the $\Delta$ interval is the difference of the two replicate distributions (conservative).}
\label{tab:app-committee-ifeval}
\end{table}

\begin{table}[h]
\centering
\setlength{\tabcolsep}{4pt}
\renewcommand{\arraystretch}{1.2}
\footnotesize
\begin{tabular}{l ccc}
\toprule
Member & HSPP individual & HSPP committee & $\Delta$ (committee $-$ individual) \\
\midrule
Gemma-27B & 1.03\spbci{0.97}{1.08} & 0.41\spbci{0.34}{0.48} & -0.62\spbci{-0.70}{-0.53} \\
Llama-Mav & 1.34\spbci{1.30}{1.38} & 0.48\spbci{0.41}{0.55} & -0.86\spbci{-0.94}{-0.78} \\
Qwen-235B & 0.83\spbci{0.62}{1.07} & 0.95\spbci{0.80}{1.11} & +0.12\spbci{-0.17}{+0.37} \\
GPT-5 & 20.15\spbci{16.57}{24.58} & 4.18\spbci{3.64}{4.69} & -15.97\spbci{-20.31}{-12.50} \\
Claude-Sonnet & 1.78\spbci{1.58}{1.99} & 1.29\spbci{1.11}{1.48} & -0.50\spbci{-0.76}{-0.20} \\
\bottomrule
\end{tabular}

\caption{LiveCodeBench: same analysis as Table~\ref{tab:app-committee-ifeval}.}
\label{tab:app-committee-lcb}
\end{table}

\paragraph{Which ensemble properties reduce self-preference?}
We enumerate all $\sum_{k=2}^{6}\binom{12}{k} = 2{,}497$ judge committees and regress the mean of the members' individual HSPP-Rub.\ Self on committee size, number of model families, and mean member MRA (Tables~\ref{tab:app-ensemble-ols-ifeval} and~\ref{tab:app-ensemble-ols-lcb}).
Judge quality (mean MRA) dominates on both benchmarks; family diversity has a small positive association on IFEval and is not significant on LiveCodeBench.

\begin{table}[h]
\centering
\setlength{\tabcolsep}{4pt}
\renewcommand{\arraystretch}{1.2}
\footnotesize
\begin{tabular}{l cc}
\toprule
Independent variable & Coef & 95\% CI \\
\midrule
Ensemble size & -0.0070 & [-0.0092, -0.0047] \\
\# families & +0.0136 & [+0.0106, +0.0166] \\
Mean MRA & +2.3279 & [+2.1669, +2.4889] \\
\bottomrule
\end{tabular}

\caption{OLS of committee-mean HSPP-Rub.\ Self on ensemble properties, IFEval ($R^2 = 0.252$).}
\label{tab:app-ensemble-ols-ifeval}
\end{table}

\begin{table}[h]
\centering
\setlength{\tabcolsep}{4pt}
\renewcommand{\arraystretch}{1.2}
\footnotesize
\begin{tabular}{l cc}
\toprule
Independent variable & Coef & 95\% CI \\
\midrule
Ensemble size & -0.0243 & [-0.1121, +0.0635] \\
\# families & +0.0475 & [-0.0687, +0.1636] \\
Mean MRA & +17.6089 & [+16.2813, +18.9365] \\
\bottomrule
\end{tabular}

\caption{OLS of committee-mean HSPP-Rub.\ Self on ensemble properties, LiveCodeBench ($R^2 = 0.227$).}
\label{tab:app-ensemble-ols-lcb}
\end{table}

\paragraph{Judge error correlation.}
Table~\ref{tab:app-krippendorff} reports the Krippendorff $\alpha$ of the five family-largest judges' rubric verdicts on subsets of increasing difficulty.
Judges agree above chance even on the subset where all generators fail ($\alpha = 0.308$ on IFEval and $\alpha = 0.154$). Since ground truth is constant on this subset, positive agreement can only reflect shared errors: judges mark the same failed rubrics as passed. These correlated false positives explain part of the self-preference that survives majority voting.

\begin{table}[h]
\centering
\setlength{\tabcolsep}{4pt}
\renewcommand{\arraystretch}{1.2}
\footnotesize
\begin{tabular}{l ccc}
\toprule
Benchmark & Rubric subset & \# items & $\alpha$ \\
\midrule
IFEval & All rubrics & 10,008 & 0.306 \\
IFEval & $\geq$1 generator failed & 4,104 & 0.314 \\
IFEval & All generators failed & 60 & 0.308 \\
LiveCodeBench & All rubrics & 69,775 & 0.353 \\
LiveCodeBench & $\geq$1 generator failed & 54,612 & 0.322 \\
LiveCodeBench & All generators failed & 1,850 & 0.154 \\
\bottomrule
\end{tabular}

\caption{Krippendorff $\alpha$ (binary, nominal) of the five family-largest judges, by rubric subset.}
\label{tab:app-krippendorff}
\end{table}

\section{Judge reasoning-effort ablation}\label{app:reasoning}

Table~\ref{tab:app-reasoning} varies the reasoning effort of the two proprietary judges on IFEval (SR).
Increased reasoning effort slightly improves MRA but does not reduce self-preference---if anything, it increases it, in line with reports that reasoning can amplify certain biases~\citep{wu2025does}.
%

\begin{table}[h]
\centering
\setlength{\tabcolsep}{4pt}
\renewcommand{\arraystretch}{1.2}
\footnotesize
\begin{tabular}{l cccc}
\toprule
Model & Reasoning & MRA & HSPP-Rub Self & HSPP-Rub Fam \\
\midrule
GPT-5 & default & 0.91\spbci{0.91}{0.92} & 1.47\spbci{0.48}{2.53} & 1.16\spbci{0.63}{1.75} \\
GPT-5 & low & 0.90\spbci{0.90}{0.91} & 1.71\spbci{0.73}{2.78} & 1.28\spbci{0.75}{1.90} \\
GPT-5 & medium & 0.91\spbci{0.91}{0.92} & 1.48\spbci{0.46}{2.55} & 1.17\spbci{0.62}{1.76} \\
GPT-5 & high & 0.92\spbci{0.91}{0.92} & 1.60\spbci{0.57}{2.82} & 1.26\spbci{0.72}{1.96} \\
Claude-Sonnet & default & 0.84\spbci{0.83}{0.85} & 0.98\spbci{0.64}{1.32} & 1.10\spbci{0.90}{1.37} \\
Claude-Sonnet & low & 0.92\spbci{0.91}{0.92} & 1.56\spbci{1.15}{2.08} & 1.40\spbci{1.12}{1.71} \\
Claude-Sonnet & medium & 0.92\spbci{0.91}{0.92} & 1.55\spbci{1.12}{2.06} & 1.46\spbci{1.16}{1.80} \\
Claude-Sonnet & high & 0.92\spbci{0.92}{0.93} & 1.63\spbci{1.14}{2.25} & 1.53\spbci{1.21}{1.94} \\
\bottomrule
\end{tabular}

\caption{IFEval (SR) rubric-level metrics by judge reasoning effort, with 95\% CIs (rubric-level bootstrap, 1000 resamples).}
\label{tab:app-reasoning}
\end{table}

\section{Generator-capability analyses}\label{app:capability}

A natural concern is that self-preference ratios are confounded by generator capability: a strong judge's own failures may be subtly wrong and thus harder to judge than the blatant failures of weaker generators.
We address this in three complementary ways; self-preference attenuates in places but never disappears.

\paragraph{Same-quality comparison sets.}
Table~\ref{tab:app-quality-tertiles} recomputes HSPP-Rub.\ Self on HealthBench using only comparator generators from the judge's own quality tertile (equal-count tertiles by reference score).

\begin{table}[h]
\centering
\setlength{\tabcolsep}{4pt}
\renewcommand{\arraystretch}{1.2}
\footnotesize
\begin{tabular}{l ccccc}
\toprule
Judge & \makecell{Score\\(as gen)} & Group & \makecell{HSPP-Rub Self\\$\cdot$ default} & \makecell{HSPP-Rub Self\\$\cdot$ same-group} & \makecell{N\\comparators} \\
\midrule
Gemma-27B & 0.586 & High & 1.10\spbci{1.08}{1.12} & 1.03\spbci{1.00}{1.05} & 3 \\
Gemma-12B & 0.572 & Mid & 1.03\spbci{1.01}{1.05} & 0.99\spbci{0.97}{1.02} & 3 \\
Gemma-4B & 0.477 & Low & 1.16\spbci{1.14}{1.17} & 1.35\spbci{1.33}{1.37} & 3 \\
Llama-Mav & 0.400 & Low & 0.71\spbci{0.69}{0.73} & 0.71\spbci{0.69}{0.74} & 2 \\
Llama-Scout & 0.392 & Low & 0.80\spbci{0.79}{0.82} & 0.82\spbci{0.80}{0.84} & 2 \\
Qwen-235B & 0.605 & High & 1.12\spbci{1.07}{1.18} & 0.95\spbci{0.90}{1.00} & 3 \\
Qwen-30B & 0.571 & Mid & 1.06\spbci{1.04}{1.08} & 1.05\spbci{1.02}{1.07} & 2 \\
Qwen-4B & 0.517 & Mid & 0.96\spbci{0.95}{0.98} & 0.98\spbci{0.96}{1.00} & 2 \\
GPT-120B & 0.630 & High & 1.00\spbci{0.94}{1.06} & 0.96\spbci{0.90}{1.02} & 2 \\
GPT-5 & 0.712 & High & 1.54\spbci{1.44}{1.64} & 1.38\spbci{1.29}{1.49} & 2 \\
Claude-Haiku & 0.500 & Low & 0.90\spbci{0.86}{0.93} & 1.13\spbci{1.09}{1.18} & 3 \\
Claude-Sonnet & 0.538 & Mid & 0.93\spbci{0.86}{1.01} & 0.96\spbci{0.87}{1.04} & 3 \\
\bottomrule
\end{tabular}

\caption{HealthBench HSPP-Rub.\ Self with the default (non-family) comparator set vs.\ only same-quality-tertile comparators. 95\% CIs from the rubric-level bootstrap (1000 resamples).}
\label{tab:app-quality-tertiles}
\end{table}

\paragraph{One-vs-one comparisons at matched capability.}
Tables~\ref{tab:app-pairwise-hb} and~\ref{tab:app-pairwise-lcb} compare GPT-5 (as judge) against single comparators of similar capability, including a variant restricted to rubrics that \emph{both} generators objectively fail, which equalizes rubric difficulty across numerator and denominator.

\begin{table}[h]
\centering
\setlength{\tabcolsep}{4pt}
\renewcommand{\arraystretch}{1.2}
\footnotesize
\begin{tabular}{l ccccc}
\toprule
Pair (GPT-5 vs \dots) & \makecell{GPT-5\\performance} & \makecell{Comparator\\performance} & \makecell{HSPP, all\\ref-negatives} & \makecell{HSPP, joint\\ref-negatives} & \makecell{R (joint\\pool)} \\
\midrule
GPT-120B & 0.71\spbci{0.71}{0.72} & 0.63\spbci{0.62}{0.64} & 1.26\spbci{1.17}{1.37} & 1.30\spbci{1.18}{1.46} & 7,479 \\
Claude-Sonnet & 0.71\spbci{0.71}{0.72} & 0.54\spbci{0.53}{0.55} & 1.52\spbci{1.40}{1.65} & 1.46\spbci{1.32}{1.62} & 8,794 \\
Qwen-235B & 0.71\spbci{0.71}{0.72} & 0.61\spbci{0.60}{0.61} & 1.33\spbci{1.23}{1.44} & 1.44\spbci{1.30}{1.59} & 8,335 \\
\bottomrule
\end{tabular}

\caption{HealthBench: GPT-5's one-vs-one self-preference against similar-capability generators. ``Joint ref-negatives'' restricts to rubrics failed by both generators (paired bootstrap); $R$ = joint pool size.}
\label{tab:app-pairwise-hb}
\end{table}

\begin{table}[h]
\centering
\setlength{\tabcolsep}{4pt}
\renewcommand{\arraystretch}{1.2}
\footnotesize
\begin{tabular}{l ccccc}
\toprule
Pair (GPT-5 vs \dots) & \makecell{GPT-5\\performance} & \makecell{Comparator\\performance} & \makecell{HSPP, all\\ref-negatives} & \makecell{HSPP, joint\\ref-negatives} & \makecell{R (joint\\pool)} \\
\midrule
GPT-120B & 0.93\spbci{0.90}{0.96} & 0.89\spbci{0.85}{0.93} & 5.49\spbci{4.03}{8.09} & 7.14\spbci{3.92}{23.52} & 221 \\
Claude-Sonnet & 0.93\spbci{0.90}{0.96} & 0.84\spbci{0.80}{0.88} & 6.23\spbci{4.68}{8.62} & 5.45\spbci{3.07}{12.80} & 209 \\
Qwen-235B & 0.93\spbci{0.90}{0.96} & 0.83\spbci{0.78}{0.88} & 19.05\spbci{12.68}{34.20} & 12.33\spbci{6.45}{39.00} & 247 \\
\bottomrule
\end{tabular}

\caption{LiveCodeBench: same analysis as Table~\ref{tab:app-pairwise-hb}.}
\label{tab:app-pairwise-lcb}
\end{table}

\paragraph{Capability predicts failure hardness.}
Table~\ref{tab:app-fail-hardness-lcb} verifies the premise of the stratified analysis of \S\ref{sec:results-objective} on LiveCodeBench: the failed completions of stronger generators concentrate in hard (low mean-MRA) cells, while those of weaker generators concentrate in easy ones.

\begin{table}[h]
\centering
\setlength{\tabcolsep}{4pt}
\renewcommand{\arraystretch}{1.2}
\footnotesize
\begin{tabular}{l ccc}
\toprule
Generator & \makecell{Pass rate} & \makecell{\% fails in\\MRA $< 0.7$ bins} & \makecell{\% fails in\\MRA $\geq 0.8$ bins} \\
\midrule
\textcolor{SwordSquash}{\large{$\boldsymbol{\cdot}$}}GPT-5 & 0.942 & 65.3\% & 21.4\% \\
\textcolor{SwordSquash}{\large{$\boldsymbol{\cdot}$}}GPT-120B & 0.900 & 68.3\% & 18.3\% \\
\textcolor{SwordPink}{\large{$\boldsymbol{\cdot}$}}Claude-Sonnet & 0.848 & 68.7\% & 8.9\% \\
\textcolor{SwordBlue}{\large{$\boldsymbol{\cdot}$}}Qwen-235B & 0.834 & 43.9\% & 24.4\% \\
\textcolor{SwordPink}{\large{$\boldsymbol{\cdot}$}}Claude-Haiku & 0.816 & 62.1\% & 10.1\% \\
\textcolor{SwordBlue}{\large{$\boldsymbol{\cdot}$}}Qwen-30B & 0.763 & 35.8\% & 27.2\% \\
\textcolor{SwordOrange}{\large{$\boldsymbol{\cdot}$}}Gemma-27B & 0.688 & 32.3\% & 30.1\% \\
\textcolor{SwordBlue}{\large{$\boldsymbol{\cdot}$}}Qwen-4B & 0.673 & 49.4\% & 27.7\% \\
\textcolor{SwordOrange}{\large{$\boldsymbol{\cdot}$}}Gemma-12B & 0.653 & 32.5\% & 37.5\% \\
\textcolor{SwordYellow}{\large{$\boldsymbol{\cdot}$}}Llama-Mav & 0.637 & 36.1\% & 17.7\% \\
\textcolor{SwordYellow}{\large{$\boldsymbol{\cdot}$}}Llama-Scout & 0.487 & 42.6\% & 20.3\% \\
\textcolor{SwordOrange}{\large{$\boldsymbol{\cdot}$}}Gemma-4B & 0.482 & 19.0\% & 55.5\% \\
\bottomrule
\end{tabular}

\caption{LiveCodeBench: distribution of each generator's failed tests across generation-hardness cells (mean rubric accuracy across all 12 judges, as in Table~\ref{tab:mra-strata-lcb}). Generator pass rate correlates negatively with the mean MRA on its completions weighted by the percentage of failed tests (Spearman $\rho = -0.71$, $p = 0.010$): better generators' mistakes are harder to judge.}
\label{tab:app-fail-hardness-lcb}
\end{table}

\paragraph{Full hardness stratification.}
Tables~\ref{tab:app-mra-ifeval}--\ref{tab:app-mra-lcb} extend the stratified analysis of \S\ref{sec:results-objective} to all twelve judges and all three benchmarks: HSPP-Rub.\ Self within bins of per-generation mean MRA across all judges (our proxy for how unambiguously good or bad a generation is; empty bins omitted).
Cells marked ``--'' have no eligible self-rubrics for that judge.

\begin{table*}[h]
\centering
\setlength{\tabcolsep}{2pt}
\renewcommand{\arraystretch}{1.2}
\scriptsize
\resizebox{\textwidth}{!}{%
\begin{tabular}{l cccccccccc}
\toprule
Judge & \makecell{0.0--0.1\\(n=44)} & \makecell{0.1--0.2\\(n=58)} & \makecell{0.2--0.3\\(n=61)} & \makecell{0.3--0.4\\(n=65)} & \makecell{0.4--0.5\\(n=109)} & \makecell{0.5--0.6\\(n=386)} & \makecell{0.6--0.7\\(n=289)} & \makecell{0.7--0.8\\(n=439)} & \makecell{0.8--0.9\\(n=768)} & \makecell{0.9--1.0\\(n=4,273)} \\
\midrule
\textcolor{SwordOrange}{\large{$\boldsymbol{\cdot}$}}Gemma-4B & 1.00\spbci{1.00}{1.00} & 1.00\spbci{1.00}{1.00} & 1.00\spbci{1.00}{1.00} & 1.00\spbci{1.00}{1.00} & 1.01\spbci{1.00}{1.05} & 1.01\spbci{1.00}{1.04} & 1.01\spbci{1.00}{1.03} & 1.04\spbci{1.01}{1.07} & 0.97\spbci{0.79}{1.10} & 1.75\spbci{0.79}{3.06} \\
\textcolor{SwordOrange}{\large{$\boldsymbol{\cdot}$}}Gemma-12B & 1.00\spbci{1.00}{1.00} & 1.00\spbci{1.00}{1.00} & 1.00\spbci{1.00}{1.00} & 1.00\spbci{1.00}{1.00} & 0.90\spbci{0.61}{1.19} & 1.01\spbci{0.91}{1.09} & 1.05\spbci{0.92}{1.15} & 0.88\spbci{0.63}{1.10} & 1.66\spbci{1.13}{2.65} & -- \\
\textcolor{SwordOrange}{\large{$\boldsymbol{\cdot}$}}Gemma-27B & 1.00\spbci{1.00}{1.00} & 1.00\spbci{1.00}{1.00} & 1.00\spbci{1.00}{1.00} & 1.00\spbci{1.00}{1.00} & 1.08\spbci{1.00}{1.22} & 1.05\spbci{1.02}{1.10} & 1.08\spbci{1.03}{1.14} & 1.11\spbci{0.91}{1.26} & 0.96\spbci{0.32}{1.61} & 0.00\spbci{0.00}{0.00} \\
\addlinespace[2pt]
\textcolor{SwordYellow}{\large{$\boldsymbol{\cdot}$}}Llama-Scout & 1.01\spbci{1.00}{1.05} & 1.03\spbci{1.00}{1.09} & 0.84\spbci{0.00}{1.21} & 0.81\spbci{0.28}{1.26} & 0.79\spbci{0.00}{1.42} & 1.11\spbci{0.91}{1.30} & 1.05\spbci{0.75}{1.33} & 0.81\spbci{0.33}{1.34} & 4.90\spbci{1.32}{12.34} & 0.00\spbci{0.00}{0.00} \\
\textcolor{SwordYellow}{\large{$\boldsymbol{\cdot}$}}Llama-Mav & 1.00\spbci{1.00}{1.00} & 1.00\spbci{1.00}{1.00} & 1.04\spbci{1.00}{1.14} & 1.08\spbci{1.02}{1.16} & 1.07\spbci{0.61}{1.41} & 1.14\spbci{0.91}{1.36} & 1.15\spbci{0.84}{1.39} & 1.15\spbci{0.55}{1.75} & 1.74\spbci{0.77}{3.74} & 0.00\spbci{0.00}{0.00} \\
\addlinespace[2pt]
\textcolor{SwordBlue}{\large{$\boldsymbol{\cdot}$}}Qwen-4B & 1.01\spbci{1.00}{1.06} & 1.01\spbci{1.00}{1.04} & 0.92\spbci{0.00}{1.41} & 1.30\spbci{1.07}{1.50} & 0.98\spbci{0.45}{1.54} & 0.94\spbci{0.74}{1.15} & 1.01\spbci{0.75}{1.26} & 1.31\spbci{0.77}{1.84} & 3.15\spbci{0.71}{7.04} & 0.00\spbci{0.00}{0.00} \\
\textcolor{SwordBlue}{\large{$\boldsymbol{\cdot}$}}Qwen-30B & 1.01\spbci{1.00}{1.05} & 1.00\spbci{1.00}{1.00} & 1.03\spbci{1.00}{1.12} & 1.04\spbci{1.00}{1.12} & 0.93\spbci{0.51}{1.21} & 0.97\spbci{0.79}{1.12} & 1.05\spbci{0.89}{1.20} & 0.98\spbci{0.63}{1.27} & 0.83\spbci{0.00}{2.17} & -- \\
\textcolor{SwordBlue}{\large{$\boldsymbol{\cdot}$}}Qwen-235B & 1.00\spbci{1.00}{1.00} & 1.03\spbci{1.00}{1.08} & -- & 1.11\spbci{1.03}{1.26} & 1.41\spbci{0.96}{2.08} & 1.34\spbci{1.05}{1.59} & 1.32\spbci{0.98}{1.64} & 1.83\spbci{1.10}{2.53} & 1.30\spbci{0.00}{3.94} & -- \\
\addlinespace[2pt]
\textcolor{SwordSquash}{\large{$\boldsymbol{\cdot}$}}GPT-120B & 1.03\spbci{1.00}{1.10} & 1.97\spbci{0.00}{5.52} & 1.27\spbci{0.00}{3.21} & 0.00\spbci{0.00}{0.00} & 4.84\spbci{3.15}{9.20} & 1.51\spbci{0.36}{2.89} & 1.91\spbci{0.90}{3.75} & 0.00\spbci{0.00}{0.00} & 0.00\spbci{0.00}{0.00} & -- \\
\textcolor{SwordSquash}{\large{$\boldsymbol{\cdot}$}}GPT-5 & 1.20\spbci{1.07}{1.46} & 0.00\spbci{0.00}{0.00} & 0.00\spbci{0.00}{0.00} & 6.27\spbci{3.41}{17.79} & -- & 1.87\spbci{0.00}{4.11} & 0.00\spbci{0.00}{0.00} & 2.56\spbci{0.00}{14.55} & 0.00\spbci{0.00}{0.00} & -- \\
\addlinespace[2pt]
\textcolor{SwordPink}{\large{$\boldsymbol{\cdot}$}}Claude-Haiku & 1.05\spbci{1.00}{1.14} & 1.15\spbci{1.05}{1.33} & 1.25\spbci{0.18}{2.20} & 0.99\spbci{0.00}{2.86} & 0.99\spbci{0.00}{3.03} & 1.11\spbci{0.74}{1.50} & 0.55\spbci{0.00}{1.43} & 0.94\spbci{0.00}{2.29} & 1.62\spbci{0.00}{9.09} & -- \\
\textcolor{SwordPink}{\large{$\boldsymbol{\cdot}$}}Claude-Sonnet & 0.82\spbci{0.00}{1.33} & 1.10\spbci{1.02}{1.26} & 1.04\spbci{0.00}{2.71} & -- & 0.57\spbci{0.00}{1.97} & 0.80\spbci{0.30}{1.35} & 1.44\spbci{0.54}{2.55} & 1.39\spbci{0.00}{4.13} & 0.00\spbci{0.00}{0.00} & -- \\
\bottomrule
\end{tabular}%
}

\caption{IFEval: HSPP-Rub.\ Self by mean-MRA bin over (generator, instance) cells; $n$ = cells per bin. 95\% CIs from the rubric-level bootstrap (1000 resamples).}
\label{tab:app-mra-ifeval}
\end{table*}

\begin{table*}[h]
\centering
\setlength{\tabcolsep}{2pt}
\renewcommand{\arraystretch}{1.2}
\scriptsize
\begin{tabular}{l ccccccc}
\toprule
Judge & \makecell{0.3--0.4\\(n=3)} & \makecell{0.4--0.5\\(n=20)} & \makecell{0.5--0.6\\(n=175)} & \makecell{0.6--0.7\\(n=1,942)} & \makecell{0.7--0.8\\(n=15,267)} & \makecell{0.8--0.9\\(n=32,064)} & \makecell{0.9--1.0\\(n=10,529)} \\
\midrule
\textcolor{SwordOrange}{\large{$\boldsymbol{\cdot}$}}Gemma-4B & 1.00\spbci{1.00}{1.00} & 0.97\spbci{0.89}{1.00} & 1.18\spbci{1.10}{1.27} & 1.07\spbci{1.03}{1.10} & 1.07\spbci{1.05}{1.09} & 1.11\spbci{1.09}{1.13} & 1.20\spbci{1.10}{1.32} \\
\textcolor{SwordOrange}{\large{$\boldsymbol{\cdot}$}}Gemma-12B & -- & 1.20\spbci{1.00}{1.60} & 0.88\spbci{0.70}{1.09} & 0.94\spbci{0.89}{1.00} & 0.96\spbci{0.93}{0.98} & 1.02\spbci{0.99}{1.04} & 1.14\spbci{1.01}{1.27} \\
\textcolor{SwordOrange}{\large{$\boldsymbol{\cdot}$}}Gemma-27B & -- & -- & 0.91\spbci{0.62}{1.18} & 1.05\spbci{0.98}{1.11} & 1.05\spbci{1.02}{1.08} & 1.10\spbci{1.06}{1.13} & 1.14\spbci{1.01}{1.29} \\
\addlinespace[2pt]
\textcolor{SwordYellow}{\large{$\boldsymbol{\cdot}$}}Llama-Scout & -- & 1.07\spbci{0.00}{2.09} & 1.09\spbci{0.87}{1.28} & 0.87\spbci{0.80}{0.95} & 0.86\spbci{0.83}{0.89} & 0.76\spbci{0.73}{0.79} & 0.71\spbci{0.62}{0.82} \\
\textcolor{SwordYellow}{\large{$\boldsymbol{\cdot}$}}Llama-Mav & -- & 1.25\spbci{0.00}{2.65} & 1.04\spbci{0.70}{1.40} & 0.83\spbci{0.75}{0.93} & 0.77\spbci{0.74}{0.80} & 0.68\spbci{0.66}{0.71} & 0.58\spbci{0.48}{0.67} \\
\addlinespace[2pt]
\textcolor{SwordBlue}{\large{$\boldsymbol{\cdot}$}}Qwen-4B & 1.00\spbci{1.00}{1.00} & -- & 0.96\spbci{0.81}{1.11} & 0.91\spbci{0.86}{0.96} & 0.95\spbci{0.93}{0.97} & 0.92\spbci{0.90}{0.95} & 0.90\spbci{0.80}{1.00} \\
\textcolor{SwordBlue}{\large{$\boldsymbol{\cdot}$}}Qwen-30B & -- & 0.68\spbci{0.00}{1.04} & 1.06\spbci{0.87}{1.23} & 1.02\spbci{0.96}{1.09} & 1.01\spbci{0.98}{1.04} & 1.06\spbci{1.03}{1.09} & 1.13\spbci{1.01}{1.25} \\
\textcolor{SwordBlue}{\large{$\boldsymbol{\cdot}$}}Qwen-235B & -- & -- & 1.92\spbci{0.43}{3.73} & 1.11\spbci{0.89}{1.37} & 1.09\spbci{1.01}{1.18} & 1.17\spbci{1.09}{1.25} & 1.23\spbci{0.98}{1.52} \\
\addlinespace[2pt]
\textcolor{SwordSquash}{\large{$\boldsymbol{\cdot}$}}GPT-120B & -- & -- & 0.32\spbci{0.00}{1.07} & 1.05\spbci{0.84}{1.27} & 1.09\spbci{1.00}{1.19} & 1.05\spbci{0.97}{1.13} & 0.83\spbci{0.63}{1.05} \\
\textcolor{SwordSquash}{\large{$\boldsymbol{\cdot}$}}GPT-5 & -- & -- & 2.79\spbci{0.49}{6.26} & 1.31\spbci{0.90}{1.78} & 1.63\spbci{1.45}{1.83} & 1.57\spbci{1.44}{1.70} & 1.80\spbci{1.40}{2.25} \\
\addlinespace[2pt]
\textcolor{SwordPink}{\large{$\boldsymbol{\cdot}$}}Claude-Haiku & -- & 3.43\spbci{0.00}{17.49} & 1.15\spbci{0.62}{1.84} & 1.08\spbci{0.94}{1.25} & 0.92\spbci{0.87}{0.98} & 0.87\spbci{0.83}{0.92} & 0.62\spbci{0.47}{0.76} \\
\textcolor{SwordPink}{\large{$\boldsymbol{\cdot}$}}Claude-Sonnet & -- & -- & 0.00\spbci{0.00}{0.00} & 1.11\spbci{0.67}{1.67} & 1.03\spbci{0.88}{1.17} & 0.90\spbci{0.80}{1.00} & 0.87\spbci{0.58}{1.17} \\
\bottomrule
\end{tabular}

\caption{HealthBench: same analysis as Table~\ref{tab:app-mra-ifeval}.}
\label{tab:app-mra-hb}
\end{table*}

\begin{table*}[h]
\centering
\setlength{\tabcolsep}{2pt}
\renewcommand{\arraystretch}{1.2}
\scriptsize
\resizebox{\textwidth}{!}{%
\begin{tabular}{l cccccccc}
\toprule
Judge & \makecell{0.2--0.3\\(n=2)} & \makecell{0.3--0.4\\(n=7)} & \makecell{0.4--0.5\\(n=29)} & \makecell{0.5--0.6\\(n=138)} & \makecell{0.6--0.7\\(n=426)} & \makecell{0.7--0.8\\(n=575)} & \makecell{0.8--0.9\\(n=513)} & \makecell{0.9--1.0\\(n=434)} \\
\midrule
\textcolor{SwordOrange}{\large{$\boldsymbol{\cdot}$}}Gemma-4B & -- & 2.04\spbci{1.31}{3.03} & 1.49\spbci{1.23}{1.79} & 2.50\spbci{1.72}{3.49} & 2.46\spbci{1.91}{3.09} & 2.25\spbci{1.75}{2.82} & 1.61\spbci{1.22}{2.15} & 0.39\spbci{0.28}{7.03} \\
\textcolor{SwordOrange}{\large{$\boldsymbol{\cdot}$}}Gemma-12B & -- & -- & 1.01\spbci{0.92}{1.09} & 1.29\spbci{1.18}{1.40} & 1.07\spbci{1.02}{1.12} & 1.11\spbci{1.03}{1.19} & 1.16\spbci{1.01}{1.31} & 1.46\spbci{1.01}{1.85} \\
\textcolor{SwordOrange}{\large{$\boldsymbol{\cdot}$}}Gemma-27B & -- & -- & 1.10\spbci{0.99}{1.18} & 1.21\spbci{1.12}{1.29} & 1.23\spbci{1.15}{1.31} & 1.02\spbci{0.92}{1.12} & 1.19\spbci{0.95}{1.49} & 5.06\spbci{2.83}{9.59} \\
\addlinespace[2pt]
\textcolor{SwordYellow}{\large{$\boldsymbol{\cdot}$}}Llama-Scout & -- & 0.97\spbci{0.91}{1.00} & -- & 1.09\spbci{1.05}{1.13} & 1.04\spbci{1.01}{1.07} & 1.28\spbci{1.23}{1.33} & 1.14\spbci{1.01}{1.27} & 1.05\spbci{0.72}{3.38} \\
\textcolor{SwordYellow}{\large{$\boldsymbol{\cdot}$}}Llama-Mav & -- & 1.02\spbci{1.00}{1.04} & 1.19\spbci{1.09}{1.27} & 1.14\spbci{1.07}{1.20} & 1.06\spbci{1.01}{1.11} & 1.53\spbci{1.46}{1.62} & 1.59\spbci{1.43}{1.76} & 0.00\spbci{0.00}{0.00} \\
\addlinespace[2pt]
\textcolor{SwordBlue}{\large{$\boldsymbol{\cdot}$}}Qwen-4B & -- & -- & 1.42\spbci{1.29}{1.55} & 2.17\spbci{1.83}{2.51} & 1.89\spbci{1.60}{2.24} & 2.29\spbci{1.81}{2.91} & 2.17\spbci{1.45}{3.24} & -- \\
\textcolor{SwordBlue}{\large{$\boldsymbol{\cdot}$}}Qwen-30B & -- & 0.74\spbci{0.47}{1.01} & 1.26\spbci{1.19}{1.32} & 1.19\spbci{0.95}{1.43} & 1.43\spbci{1.29}{1.59} & 0.95\spbci{0.78}{1.15} & 0.40\spbci{0.19}{0.67} & 0.00\spbci{0.00}{0.00} \\
\textcolor{SwordBlue}{\large{$\boldsymbol{\cdot}$}}Qwen-235B & -- & 0.67\spbci{0.36}{1.02} & -- & 0.78\spbci{0.40}{1.23} & 0.93\spbci{0.58}{1.38} & 0.25\spbci{0.06}{0.61} & 2.28\spbci{0.79}{4.49} & -- \\
\addlinespace[2pt]
\textcolor{SwordSquash}{\large{$\boldsymbol{\cdot}$}}GPT-120B & 2.66\spbci{1.71}{5.08} & -- & 8.36\spbci{5.69}{13.81} & 1.13\spbci{0.59}{1.84} & 1.98\spbci{0.83}{3.24} & 3.98\spbci{1.48}{7.33} & 7.35\spbci{1.34}{21.10} & 0.00\spbci{0.00}{0.00} \\
\textcolor{SwordSquash}{\large{$\boldsymbol{\cdot}$}}GPT-5 & -- & 8.93\spbci{5.97}{15.89} & -- & 13.74\spbci{9.94}{19.29} & 14.66\spbci{8.94}{21.48} & 7.36\spbci{1.50}{15.55} & 0.00\spbci{0.00}{0.00} & -- \\
\addlinespace[2pt]
\textcolor{SwordPink}{\large{$\boldsymbol{\cdot}$}}Claude-Haiku & -- & -- & 0.08\spbci{0.00}{0.26} & 1.05\spbci{0.84}{1.28} & 1.56\spbci{1.40}{1.76} & 2.78\spbci{2.28}{3.34} & 1.20\spbci{0.40}{2.21} & -- \\
\textcolor{SwordPink}{\large{$\boldsymbol{\cdot}$}}Claude-Sonnet & 1.00\spbci{1.00}{1.00} & -- & -- & 1.86\spbci{1.56}{2.14} & 1.33\spbci{1.11}{1.60} & 2.06\spbci{1.45}{2.89} & 2.76\spbci{0.89}{5.22} & 92.53\spbci{29.93}{6939.00} \\
\bottomrule
\end{tabular}%
}

\caption{LiveCodeBench: same analysis as Table~\ref{tab:app-mra-ifeval}.}
\label{tab:app-mra-lcb}
\end{table*}

\section{Output-property mechanisms}\label{app:mechanisms}

We test whether three output properties known to influence LLM judges explain the observed self-preference: generation length, familiarity as measured by perplexity~\citep{wataoka2024self}, and self-recognition~\citep{panickssery2024llm}.
For each property, we recompute the aggregate rubric-level HSPP-Self ratio within bins of the property, so that self and non-self outputs are compared at matched values.
Perplexity and self-recognition require access to model internals or additional prompting, so those analyses cover the nine open-weights judges; length covers all twelve.

\paragraph{Length.}
Table~\ref{tab:app-mech-length} bins each judge's own outputs by within-judge length quartile.
Self-preference grows with length on HealthBench and LiveCodeBench, consistent with longer outputs being harder to evaluate.

\begin{table}[h]
\centering
\setlength{\tabcolsep}{4pt}
\renewcommand{\arraystretch}{1.2}
\footnotesize
\begin{tabular}{l cccc}
\toprule
Benchmark & Q1 (shortest) & Q2 & Q3 & Q4 (longest) \\
\midrule
IFEval & 1.07\spbci{0.78}{1.44} & 1.22\spbci{0.97}{1.59} & 1.17\spbci{0.97}{1.43} & 1.02\spbci{0.88}{1.21} \\
HealthBench & 0.96\spbci{0.94}{0.99} & 1.01\spbci{0.98}{1.04} & 1.05\spbci{1.02}{1.08} & 1.10\spbci{1.07}{1.12} \\
LiveCodeBench & 0.90\spbci{0.79}{1.11} & 3.12\spbci{2.56}{3.89} & 2.66\spbci{2.20}{3.24} & 5.32\spbci{4.43}{6.62} \\
\bottomrule
\end{tabular}

\caption{Aggregate HSPP-Rub.\ Self by quartile of the judge's own output length (12 judges). 95\% CIs from the rubric-level bootstrap (1000 resamples).}
\label{tab:app-mech-length}
\end{table}

\paragraph{Perplexity (familiarity).}
Table~\ref{tab:app-mech-perplexity} bins every output (self and non-self) by within-judge perplexity quartile, with quartile thresholds computed per judge over all twelve generators' outputs.
Contrary to the hypothesis that controlling for perplexity eliminates self-preference, the ratio remains above 1 in the lowest-perplexity quartile on LiveCodeBench (1.48) and HealthBench (1.03); the upper quartiles are unreliable because judges' own outputs rarely land there.

\begin{table}[h]
\centering
\setlength{\tabcolsep}{4pt}
\renewcommand{\arraystretch}{1.2}
\footnotesize
\begin{tabular}{l cccc}
\toprule
Benchmark & Q1 (lowest ppl) & Q2 & Q3 & Q4 (highest ppl) \\
\midrule
IFEval & 0.94\spbci{0.89}{1.03} & 0.44\spbci{0.00}{1.42} & 3.00\spbci{0.00}{7.05} & 1.51\spbci{0.00}{5.49} \\
HealthBench & 1.03\spbci{1.01}{1.06} & 0.76\spbci{0.07}{1.45} & 0.23\spbci{0.00}{0.50} & 0.17\spbci{0.00}{0.40} \\
LiveCodeBench & 1.48\spbci{1.10}{2.61} & 0.70\spbci{0.00}{2.27} & 17.63\spbci{0.00}{44.17} & 0.00\spbci{0.00}{0.00} \\
\bottomrule
\end{tabular}

\caption{Aggregate HSPP-Rub.\ Self by within-judge perplexity quartile of the evaluated output (9 open-weights judges).}
\label{tab:app-mech-perplexity}
\end{table}

\paragraph{Self-recognition.}
We prompt each judge to predict whether each output was generated by itself, and split its own outputs by the prediction (Table~\ref{tab:app-mech-selfrecog}).
Recognition separates self-preference cleanly, especially on LiveCodeBench (1.88 when recognised vs.\ 0.72 when not), in line with~\citet{panickssery2024llm}.

\begin{table}[h]
\centering
\setlength{\tabcolsep}{4pt}
\renewcommand{\arraystretch}{1.2}
\footnotesize
\begin{tabular}{l cc}
\toprule
Benchmark & Recognised as self & Not recognised \\
\midrule
IFEval & 1.16\spbci{1.06}{1.29} & 1.09\spbci{0.99}{1.21} \\
HealthBench & 1.01\spbci{0.99}{1.03} & 0.78\spbci{0.67}{0.94} \\
LiveCodeBench & 1.88\spbci{1.75}{2.02} & 0.72\spbci{0.67}{0.78} \\
\bottomrule
\end{tabular}

\caption{Aggregate HSPP-Rub.\ Self split by the judge's self-recognition prediction on its own outputs (9 open-weights judges).}
\label{tab:app-mech-selfrecog}
\end{table}

\section{HealthBench reference ablations}\label{app:reference-ablations}

Our HealthBench reference is a fixed five-family ensemble, which is deliberately lenient: every judge's family is represented in the reference, so measured self-preference is, if anything, an underestimate.
Table~\ref{tab:app-lofo-loo} bounds this effect by rebuilding the reference as a leave-one-family-out (LOFO) or leave-one-out (LOO) majority vote for each judge.
The GPT judges move most (GPT-oss-120B: $1.00 \rightarrow 1.26/1.30$; GPT-5: $1.54 \rightarrow 1.67/1.68$), confirming that the fixed-ensemble numbers are conservative.

\begin{table}[h]
\centering
\setlength{\tabcolsep}{4pt}
\renewcommand{\arraystretch}{1.2}
\footnotesize
\begin{tabular}{l ccc}
\toprule
Judge & HSPP-Rub Self (original) & HSPP-Rub Self (LOFO) & HSPP-Rub Self (LOO) \\
\midrule
Gemma-4B & 1.16\spbci{1.14}{1.17} & 1.16\spbci{1.15}{1.18} & 1.17\spbci{1.15}{1.19} \\
Gemma-12B & 1.03\spbci{1.01}{1.05} & 1.03\spbci{1.01}{1.05} & 1.00\spbci{0.98}{1.02} \\
Gemma-27B & 1.10\spbci{1.08}{1.12} & 1.10\spbci{1.08}{1.12} & 1.09\spbci{1.07}{1.12} \\
Llama-Scout & 0.80\spbci{0.79}{0.82} & 0.79\spbci{0.77}{0.81} & 0.82\spbci{0.80}{0.84} \\
Llama-Mav & 0.71\spbci{0.69}{0.73} & 0.70\spbci{0.68}{0.72} & 0.70\spbci{0.68}{0.72} \\
Qwen-4B & 0.96\spbci{0.95}{0.98} & 0.95\spbci{0.93}{0.97} & 0.95\spbci{0.93}{0.97} \\
Qwen-30B & 1.06\spbci{1.04}{1.08} & 1.06\spbci{1.03}{1.08} & 1.07\spbci{1.04}{1.09} \\
Qwen-235B & 1.12\spbci{1.07}{1.18} & 1.14\spbci{1.10}{1.18} & 1.16\spbci{1.12}{1.21} \\
GPT-120B & 1.00\spbci{0.94}{1.06} & 1.26\spbci{1.21}{1.32} & 1.30\spbci{1.24}{1.36} \\
GPT-5 & 1.54\spbci{1.44}{1.64} & 1.67\spbci{1.60}{1.74} & 1.68\spbci{1.61}{1.76} \\
Claude-Haiku & 0.90\spbci{0.86}{0.93} & 0.90\spbci{0.86}{0.94} & 0.91\spbci{0.87}{0.94} \\
Claude-Sonnet & 0.93\spbci{0.86}{1.01} & 0.94\spbci{0.89}{1.00} & 0.90\spbci{0.85}{0.95} \\
\bottomrule
\end{tabular}

\caption{HealthBench HSPP-Rub.\ Self under the paper's ensemble reference vs.\ LOFO and LOO references. 95\% CIs from the rubric-level bootstrap (1000 resamples).}
\label{tab:app-lofo-loo}
\end{table}

\section{Mixed-effects regression details}\label{app:mixed-effects}

Figure~\ref{fig:hb-mixed-effects} summarizes a Bayesian mixed-effects logistic regression of rubric-level overestimation on HealthBench.
The unit of observation is one rubric judgment on the error-eligible population (positive rubrics the reference marks unmet, negative rubrics it marks met---the rubrics where overestimation is possible).
Fixed effects: \texttt{is\_self}, \texttt{is\_same\_family}, rubric polarity, standardized log criterion length, one dummy per axis (reference: Accuracy) and theme (reference: untagged), plus \texttt{is\_self} $\times$ (polarity + length + axis + theme) interaction terms.
Random effects: crossed random intercepts on judge and instance, which absorb each judge's baseline leniency and each instance's difficulty.
We fit with mean-field variational Bayes (\texttt{BinomialBayesMixedGLM}, statsmodels) on a 400-instance subsample ($n = 211{,}872$ judgments); variational posterior standard deviations are known to be slightly optimistic, so intervals should be read as approximate.
The interaction coefficients, shown in Figure~\ref{fig:hb-mixed-effects} and listed in Table~\ref{tab:app-mixed-interactions}, are the quantities of interest: a positive interaction means the rubric property amplifies overestimation \emph{more when the judge evaluates its own output}, i.e., it reinforces self-preference specifically.
This holds for negative polarity and for the Communication, Emergency Referrals, Global Health, and Complex Responses themes, whereas criterion length and the rubric axes show no self-specific effect.
The pooled \texttt{is\_self} main effect, by contrast, is indistinguishable from zero because self-preferring and self-deprecating judges cancel out in a single pooled term, and because the ensemble reference partially absorbs each judge's own leniency (Appendix~\ref{app:reference-ablations}).

\begin{table}[h]
\centering
\setlength{\tabcolsep}{3pt}
\renewcommand{\arraystretch}{1.2}
\footnotesize
\begin{tabular}{l cccc}
\toprule
Effect & Posterior mean & Posterior SD & 95\% interval & Odds ratio \\
\midrule
is\_self $\times$ Axis: communication\_quality & -0.033 & 0.080 & [-0.190, +0.124] & 0.968 \\
is\_self $\times$ Axis: completeness & -0.008 & 0.029 & [-0.065, +0.049] & 0.992 \\
is\_self $\times$ Axis: context\_awareness & +0.010 & 0.041 & [-0.072, +0.091] & 1.010 \\
is\_self $\times$ Axis: instruction\_following & -0.054 & 0.086 & [-0.223, +0.115] & 0.948 \\
is\_self $\times$ Theme: communication & +1.877 & 0.894 & [+0.124, +3.630] & 6.534 \\
is\_self $\times$ Theme: complex & +0.411 & 0.153 & [+0.111, +0.712] & 1.509 \\
is\_self $\times$ Theme: context & +0.255 & 0.181 & [-0.100, +0.610] & 1.291 \\
is\_self $\times$ Theme: emergency & +0.767 & 0.305 & [+0.169, +1.365] & 2.153 \\
is\_self $\times$ Theme: global & +0.761 & 0.335 & [+0.105, +1.417] & 2.140 \\
is\_self $\times$ Theme: health & +0.237 & 0.319 & [-0.388, +0.861] & 1.267 \\
is\_self $\times$ Theme: hedging & +0.110 & 0.143 & [-0.171, +0.391] & 1.116 \\
is\_self $\times$ Negative rubric & +0.081 & 0.035 & [+0.012, +0.149] & 1.084 \\
is\_self $\times$ Criterion length & -0.024 & 0.018 & [-0.060, +0.012] & 0.976 \\
\bottomrule
\end{tabular}

\caption{Interaction coefficients \texttt{is\_self} $\times$ $X$ of the HealthBench mixed-effects regression (main effects included in the model but omitted here). Positive $\Rightarrow$ property $X$ reinforces self-preference.}
\label{tab:app-mixed-interactions}
\end{table}

\section{Absolute values behind the HSPP ratios}\label{app:counts}

HSPP is a ratio; Tables~\ref{tab:app-counts-ifeval}--\ref{tab:app-counts-lcb} expose the quantities behind it for every judge and benchmark: the false-positive and eligible-rubric counts on the judge's own outputs (numerator), the pooled counts over non-family comparators, and the mean per-generator comparator false-positive rate (the actual denominator).

\begin{table*}[h]
\centering
\setlength{\tabcolsep}{4pt}
\renewcommand{\arraystretch}{1.2}
\footnotesize
\begin{tabular}{l ccccc}
\toprule
Judge & \makecell{Self FP /\\eligible} & Self FPR & \makecell{Others FP / eligible\\(pooled, non-family)} & \makecell{Mean per-gen\\other FPR} & \makecell{HSPP-Rub\\Self (point)} \\
\midrule
Gemma-4B & 141 / 143 & 0.9860 & 651 / 679 & 0.9577 & 1.03 \\
Gemma-12B & 96 / 109 & 0.8807 & 580 / 679 & 0.8553 & 1.03 \\
Gemma-27B & 94 / 99 & 0.9495 & 598 / 679 & 0.8761 & 1.08 \\
Llama-Scout & 61 / 86 & 0.7093 & 596 / 877 & 0.6771 & 1.05 \\
Llama-Mav & 53 / 67 & 0.7910 & 624 / 877 & 0.7056 & 1.12 \\
Qwen-4B & 68 / 92 & 0.7391 & 549 / 783 & 0.6971 & 1.06 \\
Qwen-30B & 75 / 90 & 0.8333 & 624 / 783 & 0.7973 & 1.05 \\
Qwen-235B & 52 / 65 & 0.8000 & 480 / 783 & 0.6166 & 1.30 \\
GPT-120B & 21 / 87 & 0.2414 & 171 / 911 & 0.1850 & 1.30 \\
GPT-5 & 7 / 32 & 0.2188 & 134 / 911 & 0.1492 & 1.47 \\
Claude-Haiku & 44 / 96 & 0.4583 & 333 / 870 & 0.3982 & 1.15 \\
Claude-Sonnet & 23 / 64 & 0.3594 & 307 / 870 & 0.3660 & 0.98 \\
\bottomrule
\end{tabular}

\caption{IFEval: absolute values and sample counts behind HSPP-Rub.\ Self.}
\label{tab:app-counts-ifeval}
\end{table*}

\begin{table*}[h]
\centering
\setlength{\tabcolsep}{4pt}
\renewcommand{\arraystretch}{1.2}
\footnotesize
\begin{tabular}{l ccccc}
\toprule
Judge & \makecell{Self FP /\\eligible} & Self FPR & \makecell{Others FP / eligible\\(pooled, non-family)} & \makecell{Mean per-gen\\other FPR} & \makecell{HSPP-Rub\\Self (point)} \\
\midrule
Gemma-4B & 12,098 / 21,563 & 0.5611 & 81,575 / 172,510 & 0.4837 & 1.16 \\
Gemma-12B & 7,040 / 17,712 & 0.3975 & 65,899 / 172,510 & 0.3870 & 1.03 \\
Gemma-27B & 6,506 / 17,128 & 0.3798 & 58,508 / 172,510 & 0.3449 & 1.10 \\
Llama-Scout & 6,330 / 25,364 & 0.2496 & 55,334 / 178,486 & 0.3110 & 0.80 \\
Llama-Mav & 4,271 / 25,063 & 0.1704 & 42,884 / 178,486 & 0.2410 & 0.71 \\
Qwen-4B & 8,455 / 19,964 & 0.4235 & 75,698 / 174,640 & 0.4397 & 0.96 \\
Qwen-30B & 7,420 / 17,839 & 0.4159 & 67,684 / 174,640 & 0.3937 & 1.06 \\
Qwen-235B & 1,683 / 16,470 & 0.1022 & 15,381 / 174,640 & 0.0909 & 1.12 \\
GPT-120B & 1,299 / 15,461 & 0.0840 & 16,842 / 201,398 & 0.0841 & 1.00 \\
GPT-5 & 998 / 12,054 & 0.0828 & 10,745 / 201,398 & 0.0539 & 1.54 \\
Claude-Haiku & 2,346 / 20,973 & 0.1119 & 22,654 / 188,618 & 0.1245 & 0.90 \\
Claude-Sonnet & 590 / 19,322 & 0.0305 & 5,880 / 188,618 & 0.0327 & 0.93 \\
\bottomrule
\end{tabular}

\caption{HealthBench: absolute values and sample counts behind HSPP-Rub.\ Self.}
\label{tab:app-counts-hb}
\end{table*}

\begin{table*}[h]
\centering
\setlength{\tabcolsep}{4pt}
\renewcommand{\arraystretch}{1.2}
\footnotesize
\begin{tabular}{l ccccc}
\toprule
Judge & \makecell{Self FP /\\eligible} & Self FPR & \makecell{Others FP / eligible\\(pooled, non-family)} & \makecell{Mean per-gen\\other FPR} & \makecell{HSPP-Rub\\Self (point)} \\
\midrule
Gemma-4B & 409 / 2,944 & 0.1389 & 1,222 / 11,959 & 0.0904 & 1.54 \\
Gemma-12B & 1,172 / 1,962 & 0.5973 & 7,469 / 11,959 & 0.6149 & 0.97 \\
Gemma-27B & 926 / 1,772 & 0.5226 & 6,318 / 11,959 & 0.5089 & 1.03 \\
Llama-Scout & 2,300 / 2,905 & 0.7917 & 8,339 / 13,637 & 0.6782 & 1.17 \\
Llama-Mav & 1,629 / 2,095 & 0.7776 & 6,692 / 13,637 & 0.5789 & 1.34 \\
Qwen-4B & 532 / 1,849 & 0.2877 & 1,561 / 14,488 & 0.1299 & 2.21 \\
Qwen-30B & 410 / 1,347 & 0.3044 & 3,568 / 14,488 & 0.2810 & 1.08 \\
Qwen-235B & 60 / 953 & 0.0630 & 748 / 14,488 & 0.0757 & 0.83 \\
GPT-120B & 115 / 578 & 0.1990 & 463 / 17,722 & 0.0333 & 5.97 \\
GPT-5 & 128 / 337 & 0.3798 & 252 / 17,722 & 0.0188 & 20.15 \\
Claude-Haiku & 396 / 1,042 & 0.3800 & 2,607 / 16,742 & 0.2227 & 1.71 \\
Claude-Sonnet & 253 / 853 & 0.2966 & 1,649 / 16,742 & 0.1662 & 1.78 \\
\bottomrule
\end{tabular}

\caption{LiveCodeBench: absolute values and sample counts behind HSPP-Rub.\ Self.}
\label{tab:app-counts-lcb}
\end{table*}

\section{Significance of the HealthBench score delta matrix}\label{app:hb-delta-matrix}

Table~\ref{tab:app-hb-delta-matrix} reports per-cell 95\% confidence intervals for the centered score delta matrix of Figure~\ref{fig:matrix-hb-sr}, from an instance-level bootstrap (1000 resamples).
Cells whose interval includes 0 are shown in gray; all remaining deltas---including every diagonal (self-evaluation) cell except Llama-Mav's and Qwen-235B's---are significant.

\begin{sidewaystable}
\centering
\setlength{\tabcolsep}{1.5pt}
\renewcommand{\arraystretch}{1.1}
\scriptsize
\begin{tabular}{l cccccccccccc}
\toprule
Judge $\downarrow$ Gen.\ $\rightarrow$ & \rotatebox{90}{Gemma-4B} & \rotatebox{90}{Gemma-12B} & \rotatebox{90}{Gemma-27B} & \rotatebox{90}{Llama-Scout} & \rotatebox{90}{Llama-Mav} & \rotatebox{90}{Qwen-4B} & \rotatebox{90}{Qwen-30B} & \rotatebox{90}{Qwen-235B} & \rotatebox{90}{GPT-120B} & \rotatebox{90}{GPT-5} & \rotatebox{90}{Claude-Haiku} & \rotatebox{90}{Claude-Sonnet} \\
\midrule
\textcolor{SwordOrange}{\large{$\boldsymbol{\cdot}$}}Gemma-4B & \makecell{+10.90\\[-1.5pt]{\tiny\textcolor{gray}{[+10.27,\,+11.55]}}} & \makecell{+8.46\\[-1.5pt]{\tiny\textcolor{gray}{[+7.86,\,+9.04]}}} & \makecell{+5.81\\[-1.5pt]{\tiny\textcolor{gray}{[+5.26,\,+6.39]}}} & \makecell{+1.55\\[-1.5pt]{\tiny\textcolor{gray}{[+0.99,\,+2.08]}}} & \textcolor{gray!60}{\makecell{+0.48\\[-1.5pt]{\tiny\textcolor{gray}{[-0.11,\,+1.06]}}}} & \makecell{+0.82\\[-1.5pt]{\tiny\textcolor{gray}{[+0.27,\,+1.39]}}} & \makecell{-2.07\\[-1.5pt]{\tiny\textcolor{gray}{[-2.58,\,-1.51]}}} & \makecell{-5.33\\[-1.5pt]{\tiny\textcolor{gray}{[-5.88,\,-4.79]}}} & \makecell{+3.85\\[-1.5pt]{\tiny\textcolor{gray}{[+3.26,\,+4.49]}}} & \makecell{-12.53\\[-1.5pt]{\tiny\textcolor{gray}{[-13.08,\,-11.90]}}} & \makecell{-4.99\\[-1.5pt]{\tiny\textcolor{gray}{[-5.60,\,-4.40]}}} & \makecell{-6.94\\[-1.5pt]{\tiny\textcolor{gray}{[-7.44,\,-6.45]}}} \\
\textcolor{SwordOrange}{\large{$\boldsymbol{\cdot}$}}Gemma-12B & \makecell{+3.05\\[-1.5pt]{\tiny\textcolor{gray}{[+2.50,\,+3.58]}}} & \makecell{-2.07\\[-1.5pt]{\tiny\textcolor{gray}{[-2.59,\,-1.55]}}} & \makecell{-2.37\\[-1.5pt]{\tiny\textcolor{gray}{[-2.88,\,-1.83]}}} & \makecell{+5.92\\[-1.5pt]{\tiny\textcolor{gray}{[+5.39,\,+6.46]}}} & \makecell{+5.42\\[-1.5pt]{\tiny\textcolor{gray}{[+4.91,\,+5.97]}}} & \makecell{+1.86\\[-1.5pt]{\tiny\textcolor{gray}{[+1.35,\,+2.42]}}} & \makecell{-0.58\\[-1.5pt]{\tiny\textcolor{gray}{[-1.02,\,-0.09]}}} & \makecell{-2.77\\[-1.5pt]{\tiny\textcolor{gray}{[-3.23,\,-2.28]}}} & \makecell{-5.72\\[-1.5pt]{\tiny\textcolor{gray}{[-6.27,\,-5.19]}}} & \makecell{-8.14\\[-1.5pt]{\tiny\textcolor{gray}{[-8.67,\,-7.63]}}} & \makecell{+3.60\\[-1.5pt]{\tiny\textcolor{gray}{[+3.07,\,+4.11]}}} & \makecell{+1.79\\[-1.5pt]{\tiny\textcolor{gray}{[+1.29,\,+2.30]}}} \\
\textcolor{SwordOrange}{\large{$\boldsymbol{\cdot}$}}Gemma-27B & \makecell{+4.40\\[-1.5pt]{\tiny\textcolor{gray}{[+3.91,\,+4.92]}}} & \makecell{-0.62\\[-1.5pt]{\tiny\textcolor{gray}{[-1.10,\,-0.22]}}} & \makecell{-1.08\\[-1.5pt]{\tiny\textcolor{gray}{[-1.52,\,-0.62]}}} & \makecell{+4.43\\[-1.5pt]{\tiny\textcolor{gray}{[+3.96,\,+4.92]}}} & \makecell{+4.11\\[-1.5pt]{\tiny\textcolor{gray}{[+3.63,\,+4.60]}}} & \makecell{+1.72\\[-1.5pt]{\tiny\textcolor{gray}{[+1.26,\,+2.19]}}} & \makecell{-0.56\\[-1.5pt]{\tiny\textcolor{gray}{[-0.97,\,-0.12]}}} & \makecell{-2.28\\[-1.5pt]{\tiny\textcolor{gray}{[-2.68,\,-1.87]}}} & \makecell{-5.20\\[-1.5pt]{\tiny\textcolor{gray}{[-5.68,\,-4.72]}}} & \makecell{-7.52\\[-1.5pt]{\tiny\textcolor{gray}{[-7.97,\,-7.03]}}} & \makecell{+2.31\\[-1.5pt]{\tiny\textcolor{gray}{[+1.84,\,+2.76]}}} & \textcolor{gray!60}{\makecell{+0.27\\[-1.5pt]{\tiny\textcolor{gray}{[-0.17,\,+0.73]}}}} \\
\addlinespace[2pt]
\textcolor{SwordYellow}{\large{$\boldsymbol{\cdot}$}}Llama-Scout & \makecell{+3.23\\[-1.5pt]{\tiny\textcolor{gray}{[+2.71,\,+3.72]}}} & \makecell{-1.06\\[-1.5pt]{\tiny\textcolor{gray}{[-1.60,\,-0.60]}}} & \makecell{-0.65\\[-1.5pt]{\tiny\textcolor{gray}{[-1.07,\,-0.19]}}} & \makecell{+2.62\\[-1.5pt]{\tiny\textcolor{gray}{[+2.16,\,+3.06]}}} & \makecell{+1.70\\[-1.5pt]{\tiny\textcolor{gray}{[+1.24,\,+2.16]}}} & \makecell{+2.51\\[-1.5pt]{\tiny\textcolor{gray}{[+2.05,\,+2.98]}}} & \makecell{+0.71\\[-1.5pt]{\tiny\textcolor{gray}{[+0.27,\,+1.16]}}} & \makecell{-0.71\\[-1.5pt]{\tiny\textcolor{gray}{[-1.13,\,-0.26]}}} & \makecell{-3.56\\[-1.5pt]{\tiny\textcolor{gray}{[-4.05,\,-3.03]}}} & \makecell{-5.22\\[-1.5pt]{\tiny\textcolor{gray}{[-5.65,\,-4.73]}}} & \makecell{+1.10\\[-1.5pt]{\tiny\textcolor{gray}{[+0.62,\,+1.57]}}} & \makecell{-0.69\\[-1.5pt]{\tiny\textcolor{gray}{[-1.16,\,-0.22]}}} \\
\textcolor{SwordYellow}{\large{$\boldsymbol{\cdot}$}}Llama-Mav & \makecell{+3.27\\[-1.5pt]{\tiny\textcolor{gray}{[+2.86,\,+3.70]}}} & \makecell{+0.66\\[-1.5pt]{\tiny\textcolor{gray}{[+0.23,\,+1.02]}}} & \textcolor{gray!60}{\makecell{+0.01\\[-1.5pt]{\tiny\textcolor{gray}{[-0.36,\,+0.39]}}}} & \makecell{+0.41\\[-1.5pt]{\tiny\textcolor{gray}{[+0.06,\,+0.83]}}} & \textcolor{gray!60}{\makecell{-0.19\\[-1.5pt]{\tiny\textcolor{gray}{[-0.57,\,+0.20]}}}} & \makecell{+1.03\\[-1.5pt]{\tiny\textcolor{gray}{[+0.68,\,+1.41]}}} & \textcolor{gray!60}{\makecell{-0.06\\[-1.5pt]{\tiny\textcolor{gray}{[-0.43,\,+0.28]}}}} & \makecell{-1.43\\[-1.5pt]{\tiny\textcolor{gray}{[-1.80,\,-1.09]}}} & \makecell{-0.91\\[-1.5pt]{\tiny\textcolor{gray}{[-1.32,\,-0.51]}}} & \makecell{-3.26\\[-1.5pt]{\tiny\textcolor{gray}{[-3.64,\,-2.89]}}} & \makecell{+0.57\\[-1.5pt]{\tiny\textcolor{gray}{[+0.17,\,+0.97]}}} & \textcolor{gray!60}{\makecell{-0.11\\[-1.5pt]{\tiny\textcolor{gray}{[-0.47,\,+0.25]}}}} \\
\addlinespace[2pt]
\textcolor{SwordBlue}{\large{$\boldsymbol{\cdot}$}}Qwen-4B & \makecell{+4.06\\[-1.5pt]{\tiny\textcolor{gray}{[+3.49,\,+4.57]}}} & \makecell{-1.93\\[-1.5pt]{\tiny\textcolor{gray}{[-2.47,\,-1.47]}}} & \makecell{-2.49\\[-1.5pt]{\tiny\textcolor{gray}{[-2.95,\,-2.02]}}} & \makecell{+6.93\\[-1.5pt]{\tiny\textcolor{gray}{[+6.39,\,+7.47]}}} & \makecell{+7.12\\[-1.5pt]{\tiny\textcolor{gray}{[+6.56,\,+7.69]}}} & \makecell{+0.88\\[-1.5pt]{\tiny\textcolor{gray}{[+0.40,\,+1.41]}}} & \makecell{-1.64\\[-1.5pt]{\tiny\textcolor{gray}{[-2.13,\,-1.13]}}} & \makecell{-2.83\\[-1.5pt]{\tiny\textcolor{gray}{[-3.28,\,-2.36]}}} & \makecell{-4.94\\[-1.5pt]{\tiny\textcolor{gray}{[-5.45,\,-4.38]}}} & \makecell{-7.44\\[-1.5pt]{\tiny\textcolor{gray}{[-7.97,\,-6.91]}}} & \makecell{+2.36\\[-1.5pt]{\tiny\textcolor{gray}{[+1.82,\,+2.91]}}} & \textcolor{gray!60}{\makecell{-0.07\\[-1.5pt]{\tiny\textcolor{gray}{[-0.56,\,+0.45]}}}} \\
\textcolor{SwordBlue}{\large{$\boldsymbol{\cdot}$}}Qwen-30B & \makecell{+4.86\\[-1.5pt]{\tiny\textcolor{gray}{[+4.35,\,+5.41]}}} & \makecell{-0.75\\[-1.5pt]{\tiny\textcolor{gray}{[-1.26,\,-0.31]}}} & \makecell{-1.33\\[-1.5pt]{\tiny\textcolor{gray}{[-1.80,\,-0.87]}}} & \makecell{+3.97\\[-1.5pt]{\tiny\textcolor{gray}{[+3.49,\,+4.47]}}} & \makecell{+3.37\\[-1.5pt]{\tiny\textcolor{gray}{[+2.88,\,+3.89]}}} & \makecell{+2.44\\[-1.5pt]{\tiny\textcolor{gray}{[+1.95,\,+2.92]}}} & \makecell{-0.49\\[-1.5pt]{\tiny\textcolor{gray}{[-0.94,\,-0.04]}}} & \makecell{-2.27\\[-1.5pt]{\tiny\textcolor{gray}{[-2.70,\,-1.84]}}} & \makecell{-4.36\\[-1.5pt]{\tiny\textcolor{gray}{[-4.84,\,-3.89]}}} & \makecell{-6.37\\[-1.5pt]{\tiny\textcolor{gray}{[-6.85,\,-5.91]}}} & \makecell{+1.58\\[-1.5pt]{\tiny\textcolor{gray}{[+1.02,\,+2.10]}}} & \makecell{-0.65\\[-1.5pt]{\tiny\textcolor{gray}{[-1.12,\,-0.17]}}} \\
\textcolor{SwordBlue}{\large{$\boldsymbol{\cdot}$}}Qwen-235B & \makecell{-0.45\\[-1.5pt]{\tiny\textcolor{gray}{[-0.76,\,-0.15]}}} & \textcolor{gray!60}{\makecell{-0.28\\[-1.5pt]{\tiny\textcolor{gray}{[-0.59,\,+0.01]}}}} & \textcolor{gray!60}{\makecell{-0.17\\[-1.5pt]{\tiny\textcolor{gray}{[-0.47,\,+0.12]}}}} & \makecell{+0.43\\[-1.5pt]{\tiny\textcolor{gray}{[+0.15,\,+0.73]}}} & \makecell{+0.67\\[-1.5pt]{\tiny\textcolor{gray}{[+0.38,\,+0.99]}}} & \textcolor{gray!60}{\makecell{+0.19\\[-1.5pt]{\tiny\textcolor{gray}{[-0.10,\,+0.48]}}}} & \makecell{+0.35\\[-1.5pt]{\tiny\textcolor{gray}{[+0.03,\,+0.66]}}} & \textcolor{gray!60}{\makecell{+0.22\\[-1.5pt]{\tiny\textcolor{gray}{[-0.07,\,+0.51]}}}} & \textcolor{gray!60}{\makecell{-0.05\\[-1.5pt]{\tiny\textcolor{gray}{[-0.33,\,+0.26]}}}} & \makecell{-0.35\\[-1.5pt]{\tiny\textcolor{gray}{[-0.64,\,-0.02]}}} & \makecell{-0.29\\[-1.5pt]{\tiny\textcolor{gray}{[-0.59,\,-0.01]}}} & \textcolor{gray!60}{\makecell{-0.26\\[-1.5pt]{\tiny\textcolor{gray}{[-0.55,\,+0.01]}}}} \\
\addlinespace[2pt]
\textcolor{SwordSquash}{\large{$\boldsymbol{\cdot}$}}GPT-120B & \makecell{-0.77\\[-1.5pt]{\tiny\textcolor{gray}{[-1.22,\,-0.32]}}} & \makecell{-1.59\\[-1.5pt]{\tiny\textcolor{gray}{[-2.04,\,-1.16]}}} & \makecell{-1.75\\[-1.5pt]{\tiny\textcolor{gray}{[-2.20,\,-1.33]}}} & \makecell{+2.77\\[-1.5pt]{\tiny\textcolor{gray}{[+2.33,\,+3.22]}}} & \makecell{+3.22\\[-1.5pt]{\tiny\textcolor{gray}{[+2.79,\,+3.66]}}} & \textcolor{gray!60}{\makecell{+0.38\\[-1.5pt]{\tiny\textcolor{gray}{[-0.05,\,+0.81]}}}} & \textcolor{gray!60}{\makecell{-0.16\\[-1.5pt]{\tiny\textcolor{gray}{[-0.55,\,+0.26]}}}} & \textcolor{gray!60}{\makecell{-0.32\\[-1.5pt]{\tiny\textcolor{gray}{[-0.70,\,+0.05]}}}} & \makecell{-1.05\\[-1.5pt]{\tiny\textcolor{gray}{[-1.50,\,-0.63]}}} & \makecell{-1.96\\[-1.5pt]{\tiny\textcolor{gray}{[-2.42,\,-1.48]}}} & \makecell{+0.93\\[-1.5pt]{\tiny\textcolor{gray}{[+0.52,\,+1.32]}}} & \textcolor{gray!60}{\makecell{+0.31\\[-1.5pt]{\tiny\textcolor{gray}{[-0.08,\,+0.71]}}}} \\
\textcolor{SwordSquash}{\large{$\boldsymbol{\cdot}$}}GPT-5 & \makecell{-2.11\\[-1.5pt]{\tiny\textcolor{gray}{[-2.59,\,-1.64]}}} & \makecell{-2.49\\[-1.5pt]{\tiny\textcolor{gray}{[-2.95,\,-2.06]}}} & \makecell{-2.63\\[-1.5pt]{\tiny\textcolor{gray}{[-3.11,\,-2.17]}}} & \makecell{+0.72\\[-1.5pt]{\tiny\textcolor{gray}{[+0.26,\,+1.21]}}} & \makecell{+2.49\\[-1.5pt]{\tiny\textcolor{gray}{[+2.08,\,+2.91]}}} & \makecell{-2.29\\[-1.5pt]{\tiny\textcolor{gray}{[-2.75,\,-1.87]}}} & \makecell{-0.75\\[-1.5pt]{\tiny\textcolor{gray}{[-1.15,\,-0.34]}}} & \makecell{+1.05\\[-1.5pt]{\tiny\textcolor{gray}{[+0.59,\,+1.47]}}} & \makecell{-1.52\\[-1.5pt]{\tiny\textcolor{gray}{[-2.00,\,-1.02]}}} & \makecell{+4.63\\[-1.5pt]{\tiny\textcolor{gray}{[+4.13,\,+5.16]}}} & \makecell{+0.88\\[-1.5pt]{\tiny\textcolor{gray}{[+0.43,\,+1.30]}}} & \makecell{+2.01\\[-1.5pt]{\tiny\textcolor{gray}{[+1.58,\,+2.45]}}} \\
\addlinespace[2pt]
\textcolor{SwordPink}{\large{$\boldsymbol{\cdot}$}}Claude-Haiku & \textcolor{gray!60}{\makecell{-0.17\\[-1.5pt]{\tiny\textcolor{gray}{[-0.63,\,+0.30]}}}} & \makecell{-0.64\\[-1.5pt]{\tiny\textcolor{gray}{[-1.10,\,-0.17]}}} & \makecell{-0.83\\[-1.5pt]{\tiny\textcolor{gray}{[-1.30,\,-0.41]}}} & \makecell{-1.70\\[-1.5pt]{\tiny\textcolor{gray}{[-2.21,\,-1.17]}}} & \makecell{-1.18\\[-1.5pt]{\tiny\textcolor{gray}{[-1.63,\,-0.68]}}} & \textcolor{gray!60}{\makecell{-0.13\\[-1.5pt]{\tiny\textcolor{gray}{[-0.59,\,+0.30]}}}} & \makecell{+0.82\\[-1.5pt]{\tiny\textcolor{gray}{[+0.35,\,+1.27]}}} & \makecell{+0.58\\[-1.5pt]{\tiny\textcolor{gray}{[+0.11,\,+1.02]}}} & \makecell{+0.85\\[-1.5pt]{\tiny\textcolor{gray}{[+0.40,\,+1.35]}}} & \makecell{+0.97\\[-1.5pt]{\tiny\textcolor{gray}{[+0.52,\,+1.48]}}} & \makecell{+0.96\\[-1.5pt]{\tiny\textcolor{gray}{[+0.51,\,+1.41]}}} & \textcolor{gray!60}{\makecell{+0.47\\[-1.5pt]{\tiny\textcolor{gray}{[-0.00,\,+0.95]}}}} \\
\textcolor{SwordPink}{\large{$\boldsymbol{\cdot}$}}Claude-Sonnet & \makecell{-1.47\\[-1.5pt]{\tiny\textcolor{gray}{[-1.90,\,-1.03]}}} & \textcolor{gray!60}{\makecell{+0.06\\[-1.5pt]{\tiny\textcolor{gray}{[-0.37,\,+0.48]}}}} & \textcolor{gray!60}{\makecell{-0.08\\[-1.5pt]{\tiny\textcolor{gray}{[-0.47,\,+0.32]}}}} & \textcolor{gray!60}{\makecell{+0.02\\[-1.5pt]{\tiny\textcolor{gray}{[-0.40,\,+0.46]}}}} & \makecell{+1.39\\[-1.5pt]{\tiny\textcolor{gray}{[+1.00,\,+1.81]}}} & \makecell{-3.87\\[-1.5pt]{\tiny\textcolor{gray}{[-4.32,\,-3.42]}}} & \makecell{-0.83\\[-1.5pt]{\tiny\textcolor{gray}{[-1.23,\,-0.42]}}} & \makecell{+1.09\\[-1.5pt]{\tiny\textcolor{gray}{[+0.70,\,+1.47]}}} & \textcolor{gray!60}{\makecell{-0.32\\[-1.5pt]{\tiny\textcolor{gray}{[-0.74,\,+0.11]}}}} & \makecell{+1.87\\[-1.5pt]{\tiny\textcolor{gray}{[+1.47,\,+2.30]}}} & \makecell{+0.61\\[-1.5pt]{\tiny\textcolor{gray}{[+0.21,\,+1.01]}}} & \makecell{+1.52\\[-1.5pt]{\tiny\textcolor{gray}{[+1.10,\,+1.92]}}} \\
\bottomrule
\end{tabular}

\caption{Per-cell 95\% confidence intervals for the centered score delta matrix of Figure~\ref{fig:matrix-hb-sr} (HealthBench, SR, weighted scoring, scaled by $\times$100): each cell shows judge $j$'s deviation for generator $g$, centered by judge $j$'s average bias, with instance-level bootstrap CIs (1000 resamples). Gray cells: the CI includes 0 (not significant).}
\label{tab:app-hb-delta-matrix}
\end{sidewaystable}

\end{document}